\documentclass[10pt,twocolumn,letterpaper]{article}

\usepackage[pagenumbers]{cvpr} % To force page numbers, e.g. for an arXiv version

\usepackage{url}
\usepackage{lineno}
\usepackage{inconsolata}
\usepackage{CJKutf8}
\usepackage{amsmath}
\usepackage{pifont}

\usepackage{subcaption}
\usepackage{booktabs, multirow}
\usepackage{soul}
\usepackage{graphicx}
\usepackage{enumitem}
\usepackage{wrapfig}
\usepackage[linesnumbered,ruled,vlined]{algorithm2e}

\usepackage[dvipsnames]{xcolor} 
\usepackage{float}      % Might be needed for [H] specifier if issues arise

\usepackage{natbib}  % DO NOT CHANGE THIS AND DO NOT ADD ANY OPTIONS TO IT
\usepackage{caption} % DO NOT CHANGE THIS AND DO NOT ADD ANY OPTIONS TO IT
\usepackage{colortbl}
\usepackage{minted}

\definecolor{darkblue}{rgb}{0, 0, 0.5}
\definecolor{darkred}{RGB}{255, 150, 113}
\definecolor{darkgreen}{RGB}{50, 206, 50}
\definecolor{lightblue}{RGB}{173, 216, 230} % 浅蓝色定义

\definecolor{cvprblue}{rgb}{0.21,0.49,0.74}
\usepackage[pagebackref,breaklinks,colorlinks,allcolors=cvprblue]{hyperref}

\def\paperID{22012} % *** Enter the Paper ID here
\def\confName{CVPR}
\def\confYear{2026}

\title{MetaphorStar: Image Metaphor Understanding and Reasoning with End-to-End Visual Reinforcement Learning}

\author{
    Chenhao Zhang\textsuperscript{1,2} \quad
    Yazhe Niu\textsuperscript{1,3} \quad
    Hongsheng Li\textsuperscript{3} \\
    \textsuperscript{1}Shanghai AI Laboratory \quad
    \textsuperscript{2}Huazhong University of Science and Technology \\
    \textsuperscript{3}The Chinese University of Hong Kong MMLab \\
    {\tt\small zhangchenhao@pjlab.org.cn} \quad {\tt\small niuyazhe314@outlook.com}
}
\begin{document}

\maketitle

\begin{abstract}
Metaphorical comprehension in images remains a critical challenge for Nowadays AI systems. While Multimodal Large Language Models (MLLMs) excel at basic Visual Question Answering (VQA), they consistently struggle to grasp the nuanced cultural, emotional, and contextual implications embedded in visual content. This difficulty stems from the task's demand for sophisticated multi-hop reasoning, cultural context, and Theory of Mind (ToM) capabilities, which current models lack. To fill this gap, we propose \textbf{MetaphorStar}, the first end-to-end visual reinforcement learning (RL) framework for image implication tasks. Our framework includes three core components: the fine-grained dataset TFQ-Data, the visual RL method TFQ-GRPO, and the well-structured benchmark TFQ-Bench.

Our fully open-source MetaphorStar family, trained using TFQ-GRPO on TFQ-Data, significantly improves performance by an average of 82.6\% on the image implication benchmarks. Compared with 20+ mainstream MLLMs, MetaphorStar-32B achieves state-of-the-art (SOTA) on Multiple-Choice Question and Open-Style Question, significantly outperforms the top closed-source model Gemini-3.0-pro on True-False Question. Crucially, our experiments reveal that learning image implication tasks improves the general understanding ability, especially the complex visual reasoning ability. We further provide a systematic analysis of model parameter scaling, training data scaling, and the impact of different model architectures and training strategies, demonstrating the broad applicability of our method.
We open-sourced all model weights, datasets and method code at \url{https://metaphorstar.github.io}.

\end{abstract}

\section{Introduction}
\label{sec:intro}

\begin{quote}
We don't see things as they are, we see them as we are.

\hfill--- Anaïs Nin
\end{quote}

% Metaphors are not just abstract concepts found in literature; they are also prevalent in our daily lives. 
% For instance, when we say ``time is money'' or ``life is a journey'', we are using metaphors to convey complex ideas in a more contextual and understandable way. 
% These metaphors highlight the integral role that metaphoric thinking plays in human communication. 
% Just as we use metaphors to make sense of the world around us, we aim to enable AI to understand metaphors in a human-like manner.
% As established in ``Metaphors We Live By''~\citep{lakoff2008metaphors}, metaphors are not merely ornamental language devices but fundamental cognitive tools that allow us to conceptualize our surroundings.
% Metaphors possess characteristics such as systematicity, the creation of similarity, and imaginative rationality. 
% Through cross-domain mapping, one concept can be used to comprehend another, allowing for a more insightful interpretation.

This sentiment captures the core challenge of this paper: the profound gap between literal perception and conceptual understanding. The quote presents a dichotomy. ``Seeing things as they are" is the realm of literal perception—the ability to identify objects and describe a scene, a task at which modern Multimodal Large Language Models (MLLMs) excel. ``Seeing things as we are," however, is the realm of implication. It means interpreting that scene through the lens of human context, culture, and shared knowledge. This is the gap where MLLMs fail.

\begin{figure}[th]
    \centering
	\includegraphics[width=0.95\linewidth]{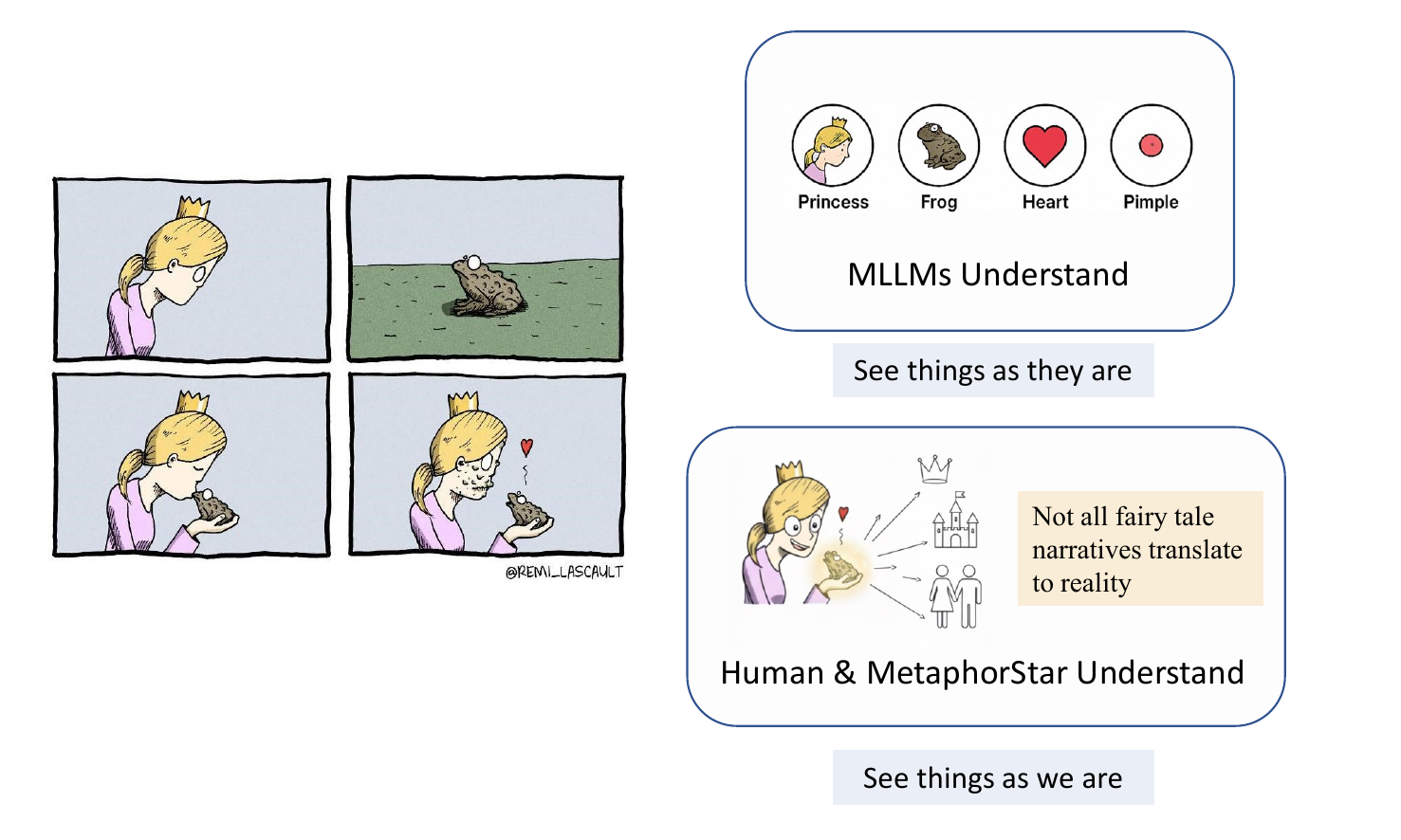}
    \vspace{-10pt}
    \caption{A picture is worth a thousand words: While MLLMs excel at literal object recognition (\textbf{``See things as they are"}), they often miss the deeper implication. Humans and our MetaphorStar model interpret the world \textbf{``See things as we are"}, grasping complex implications which behind the simple factual descriptions.}
    \vspace{-10pt}
    \label{fig:teaser}
\end{figure}

This gap is the essence of metaphorical comprehension, as visualized in Figure~\ref{fig:teaser}. Metaphors are not just abstract concepts found in literature, such as ``time is money'' or ``life is a journey,'' but fundamental cognitive tools that allow us to conceptualize our surroundings~\citep{lakoff2008metaphors}. In our daily lives, we are surrounded by \textit{visual} metaphors: a political cartoon depicting a government as a ``ship of state," an image of a person literally ``at a crossroads," or a ``wilted plant" on an office desk. An MLLM might see ``a person" and ``a split road" (seeing as it is), but it fails to infer the implication of a ``life-changing decision" (seeing as we are). These images convey complex ideas by mapping one conceptual domain onto another. Just as humans use this abstract thinking to make sense of the world, we aim to enable AI to bridge this gap and truly understand these implications.

% With the rapid advancement of large language models (LLMs), models such as OpenAI o1 \citep{o1} and DeepSeek-R1 \citep{deepseek-R1} have demonstrated remarkable text-reasoning capabilities. 
% However, a significant amount of knowledge in the real world cannot be fully represented by text alone. 
% Visual information, for instance, contains a wealth of knowledge that is not easily captured through text. 
% As a result, there has been a growing interest in integrating visual information into text-reasoning tasks. 
% Compared to language, vision is inherently complex due to its diverse representation, subjective understanding, and difficulty in quantifying its data.

In recent years, vision-language reasoning models such as o3 \citep{o4-mini}, Gemini-2.5-pro \citep{gemini2}, and Grok-3-reasoning \citep{grok3} have achieved outstanding performance. 
For example, Gemini-2.5-pro has reached a high score on math, code and vision-language reasoning benchmarks \citep{lightman2023letsverifystepstep,lu2024mathvista,wang2024measuring,yue2023mmmu}. 
However, these models still struggle with image metaphor questions \citep{liu2024iibenchimageimplicationunderstanding,zhang2024mllmsunderstanddeepimplication}.
They tend to focus on the superficial elements of the image, neglecting the deeper connections and emotional expressions among them.  
It is important to note that these models excel at logical reasoning tasks, which are based on a different set of cognitive principles compared to image metaphor. 
Unlike VQA tasks that focus on concrete image comprehension, image metaphors require a stronger emphasis on abstract meaning and higher-order reasoning abilities.
It is not a simple logical reasoning task and needs a different method to understand implications.
It requires the model to grasp complex and abstract information, such as metaphors, symbols, and emotions in the image, rather than just concrete contents.

Understanding image implication is a more complex and challenging task than conventional VQA tasks. 
It requires advanced cognitive abilities such as multi-hop reasoning and a sophisticated theory of mind (ToM), which are inherent to human cognition \citep{liu2024iibenchimageimplicationunderstanding,zhang2024mllmsunderstanddeepimplication}. 

% Existing methods for image metaphor understanding mainly fall into three categories: explicit metaphor mapping, model implicit reasoning, and contextual alignment. Explicit mapping method creates a link between metaphor ontology and visual representation. For example, the CLOT method \citep{zhong2024lets} uses this mapping to understand image metaphors. However, it struggles with complex many-to-many mappings and dynamically changing cultural backgrounds. Implicit reasoning method relies on the model’s ability to reason without explicit mapping. For instance, the C4MMD method \citep{xu-etal-2024-exploring} uses a training-free reasoning approach. Despite its potential, it still faces challenges in handling complex metaphor understanding tasks without training model, especially in situations involving multimodal information and cultural backgrounds. Contextual alignment method uses out-of-domain knowledge to align with the dynamically changing cultural image metaphors. Despite its innovative and effective approach, this strategy has significant drawbacks. Its reliance on external knowledge sources, such as the internet, introduces unpredictability, as the quality and relevance of retrieved information are not guaranteed. Furthermore, the iterative process of querying models and performing external searches is computationally intensive, consuming considerable time and resources.

Existing methods for image metaphor understanding mainly fall into three categories. (1) Explicit mapping, represented by CLOT \citep{zhong2024lets}, creates links between metaphor ontologies and visual representations. It struggles with complex many-to-many mappings and dynamic cultural references. (2) {Implicit reasoning}, exemplified by C4MMD \citep{xu-etal-2024-exploring}, uses training-free CoT reasoning. This passive approach often fails to handle the complex search space of abstract thought. (3) {Contextual alignment} \citep{zhang2025letandroidsdream} uses out-of-domain knowledge to align with cultural metaphors. This strategy introduces unpredictability from external knowledge quality and is computationally intensive.

To address these problems, we posit that a new approach is needed. Inspired by human cognitive models like the DIKW pyramid \citep{dikw}, we believe that the implicit reasoning capabilities within MLLMs should be sufficient, but they remain dormant, lacking a method to effectively activate this latent knowledge. Passive, training-free CoT prompting is often too weak to ``find" the correct reasoning path. In contrast, Reinforcement Learning (RL) provides an active mechanism to explicitly reward and reinforce the model for exploring and strengthening these complex, non-literal reasoning pathways.

Therefore, we propose \textbf{MetaphorStar}, the first end-to-end visual RL framework for image implication.
Our framework includes three core components: the fine-grained dataset {TFQ-Data}, the visual RL method {TFQ-GRPO}, and the well-structured benchmark {TFQ-Bench}.
Our open-sourced MetaphorStar family, trained using this method, achieves state-of-the-art performance, and experiments consistently verify its superiority across TFQ, MCQ, and OSQ formats.
\textit{Our contributions are listed as follows:}

\begin{itemize}[leftmargin=*]
\item 

We systematically analyze the image implication task and find that learning it helps improve general understanding ability, especially the complex visual reasoning ability, as demonstrated through sufficient experiments.
\item 
To the best of our knowledge, we propose the first end-to-end RL framework for image implication tasks, including the fine-grained dataset TFQ-Data, the visual RL method, and the well-structured benchmark TFQ-Bench.
\item 
Our fully open-scoured MetaphorStar family, trained using TFQ-GRPO on TFQ-Data, significantly improves performance by an average of 82.6\% on the image implication benchmark. Compared with 20+ mainstream MLLMs, MetaphorStar-32B achieves SOTA on Multiple-Choice Question and Open-Style Question, significantly outperforms the top closed-source model Gemini-3.0-pro on True-False Question, and generalizes well on general VQA tasks.
\end{itemize}

\section{Related Work}
\label{sec:related work}

\subsection{Image Implication}

Image implication encompasses diverse cognitive phenomena, including humor, sarcasm, and broader metaphorical understanding. Early research in this domain often focused on specific aspects, such as humor recognition \citep{hessel-etal-2023-androids,horvitz-etal-2024-getting} and sarcasm detection \citep{Desai_Chakraborty_Akhtar_2022}. The rapid development of Large Language Models (LLMs) presents new opportunities for analyzing these implications, necessitating more comprehensive evaluation frameworks. To this end, DeepEval \citep{yang-etal-2024-large} provided a systematic taxonomy of image implications. Subsequently, II-Bench \citep{liu2024iibenchimageimplicationunderstanding} emerged as the first English image implication benchmark, followed by CII-Bench \citep{zhang2024mllmsunderstanddeepimplication}, which extended this framework to Chinese images.

Image implication understanding requires sophisticated multi-hop reasoning and theory of mind (ToM) capabilities \citep{liu2024iibenchimageimplicationunderstanding,zhang2024mllmsunderstanddeepimplication}. Current methods generally fall into three categories.
First, {explicit metaphor mapping} (e.g., CLOT \citep{zhong2024lets}) links visual features to metaphor ontologies. This approach is limited by the complexity of many-to-many metaphorical relationships and the static nature of ontologies, which fail to capture dynamic cultural references.
Second, {model implicit reasoning} (e.g., C4MMD \citep{xu-etal-2024-exploring}) utilizes techniques like Chain-of-Thought (CoT) prompting. However, it struggles with the non-logical nature of metaphor and the vast search space required for out-of-domain reasoning.
Third, {contextual alignment} (e.g., LAD \citep{zhang2025letandroidsdream}) iteratively enriches image captions with knowledge from external sources. This strategy is computationally intensive and hindered by the unreliable quality of retrieved external information.

\subsection{Vision-language Reasoning}

The rapid advancement of LLMs has demonstrated remarkable text reasoning capabilities, as evidenced by models such as o1 \citep{o1} and DeepSeek-R1 \citep{deepseek-R1}.
However, real-world knowledge often transcends textual representation, with visual information encapsulating world knowledge that pure language models cannot access.
For example, images inherently contain rich, multi-layered information that often resists straightforward textual description, including spatial relationships, contextual nuances, and implicit knowledge that humans process intuitively.  
This limitation has driven research toward integrating visual information into reasoning frameworks. 
Current research has developed three primary approaches to incorporate visual information into model reasoning:
1) Comprehensive MLLM Description: This approach treats visual content as the text grounding problem, as demonstrated by LLAVA-COT \citep{xu2024llavacotletvisionlanguage} and Mulberry \citep{yao2024mulberryempoweringmllmo1like}. 
2) Multi-turn MLLM Interaction: Models like VoCoT \citep{li2024vocotunleashingvisuallygrounded}, V* \citep{wu2024v}, o3 \citep{o4-mini}, Gemini-2.5-pro \citep{gemini2}, and DeepEyes \citep{zheng2025deepeyesincentivizingthinkingimages} employ iterative question-answering to extract fine-grained visual information at various levels of detail.
3) Tool-augmented Reasoning: Frameworks such as Visual Sketchpad \citep{hu2024visual}, Whiteboard-of-Thought \citep{menon2024whiteboard}, o3 \citep{o4-mini}, Gemini-2.5-pro \citep{gemini2}, and DeepEyes \citep{zheng2025deepeyesincentivizingthinkingimages} leverage tool-based approaches to modify images and augment reasoning with prior knowledge embedded in these tools.

\section{Method}
\label{sec:method}
\subsection{True-False Question (TFQ) For Image Implication Understanding}

Previous benchmarks have advanced the evaluation of image implication understanding through diverse question formats. II-Bench \citep{liu2024iibenchimageimplicationunderstanding} introduced the Multiple-Choice Question (MCQ), which offers a balanced assessment of a model's comprehension. Subsequently, CII-Bench \citep{zhang2024mllmsunderstanddeepimplication} proposed the Open-Style Question (OSQ), which represents an upper bound on task difficulty due to its high degree of openness and the sophisticated reasoning it demands.

Our analysis of these formats reveals a clear spectrum of challenges. While MCQ provides a stable, medium-difficulty evaluation and OSQ tests the limits of generative reasoning, there is a need for a more foundational and comprehensive assessment tool. To fill this gap, we introduce the True-False Question (TFQ). The TFQ task is designed as a fine-grained complement to MCQ, establishing a lower bound on difficulty. Unlike formats that target a single inferential conclusion, TFQ probes understanding across multiple dimensions by presenting a series of statements about an image. These statements cover not only the central implication but also essential visual information, akin to basic VQA, thereby ensuring a more holistic evaluation of a model's capabilities from perception to cognition.

As summarized in Table~\ref{tab:question_comparison}, the three formats offer a complementary suite for evaluation. We analyze them across three key dimensions essential for Reinforcement Learning:
\begin{itemize}
    \item \textbf{Knowledge Density:} The breadth of factual and inferential points evaluated per image. TFQ ranks highest as it forces the model to verify multiple distinct propositions per image.
    \item \textbf{Learnability:} The ease with which a model can learn from the signal. TFQ provides a clearer, less noisy gradient signal compared to the complex search space of OSQ.
    \item \textbf{Verifiability:} The objectivity of the ground truth. TFQ offers definitive binary answers, avoiding the subjective ambiguity of open-ended generation.
\end{itemize}
This makes TFQ the ideal substrate for our visual RL framework, providing a dense and verifiable reward signal.

\begin{table}[htb]
\centering
\scalebox{0.9}{
\setlength{\tabcolsep}{10pt}
\begin{tabular}{lccc} 
\toprule
\textbf{Ability} & \textbf{TFQ} & \textbf{MCQ} & \textbf{OSQ} \\ 
\midrule
% Difficulty & $\star$ & $\star\star$ & $\star\star\star$ \\
% Openness & $\star$ & $\star\star$ & $\star\star\star$ \\
Knowledge Density & $\star\star\star$ & $\star\star$ & $\star$ \\
Learnability & $\star\star\star$ & $\star\star$ & $\star$ \\
Verifiability & $\star\star\star$ & $\star\star$ & $\star$ \\
\bottomrule
\end{tabular}
}
\caption{Comparison of True-False Question, Multiple-Choice Question, and Open-Style Question across different dimensions. TFQ offers superior properties for training, including higher knowledge density, better learnability and higher verifiability. Relative ranking: $\star\star\star$ (Highest) $>$ $\star\star$ (Medium) $>$ $\star$ (Lowest).}
\label{tab:question_comparison}
\vspace{-10pt}
\end{table}

\subsection{TFQ-Data \& TFQ-Bench}

\subsubsection{Data Generation}
To construct our dataset, we leveraged the 1,434 high-quality metaphorical images from the II-Bench \citep{liu2024iibenchimageimplicationunderstanding}. We utilized the GPT-4.1 model to generate a comprehensive set of TFQs. For each image, the model was provided with its detailed textual description and the ground-truth implication, prompting it to generate an average of 5-10 QA pairs, each with a definitive True/False answer. This process yields a total collection of 14,099 questions. We also {manually verify} the generated data.

The question design was guided by several principles to ensure comprehensiveness. First, each TFQ is a proposition that evaluates understanding of key image content related to the central metaphor. Second, the questions are not confined to the implication itself but also probe the model's grasp of primary visual information (akin to basic VQA). Third, the set of questions for each image includes hierarchical difficulty levels; false statements are crafted to be plausible distractors, while true statements are clearly grounded in the visual or contextual evidence.

\subsubsection{Dataset and Benchmark Splits}
We partition the total collection (1,434 images, 14,099 questions) into dedicated sets for training (TFQ-Data) and evaluation (TFQ-Bench), as summarized in Table~\ref{tab:dataset_splits}. The detailed statistic is in Appendix~\ref{app:statistic}. 

\textbf{TFQ-Data.} The training set is provided in two scales. {TFQ-Data-Full} is the large-scale training set, containing 1,384 images and 13,607 questions. From this set, we also curate TFQ-Data-Lite, a smaller (100 images, 984 questions) subset hand-picked for its high quality, diversity, and richness, making it ideal for rapid experimentation.

\textbf{TFQ-Bench.} The evaluation component also exists at two scales. TFQ-Bench-Full refers to the entire dataset (1,434 images, 14,099 questions). TFQ-Bench-Lite is the efficient test set, containing 50 representative images and 492 questions, used for concise and standardized evaluation. Crucially, this TFQ-Bench-Lite set is strictly disjoint from the TFQ-Data-Full training set, ensuring a fair and rigorous evaluation of model performance.

\begin{table}[ht]
\centering
\scalebox{0.7}{
\setlength{\tabcolsep}{10pt}
\begin{tabular}{lcccc}
\toprule
\textbf{Type} & \textbf{Split} & \textbf{Purpose} & \textbf{Images} & \textbf{Questions} \\
\midrule
\multirow{2}{*}{\textbf{TFQ-Data}} & \textbf{Lite} & Efficient Fine-tuning & 100 & 984 \\
& \textbf{Full} & Large-scale Training & 1,384 & 13,607 \\
\midrule
\multirow{2}{*}{\textbf{TFQ-Bench}} & \textbf{Lite} & Efficient Evaluate & 50 & 492 \\
& \textbf{Full} & Full Benchmark & 1,434 & 14,099 \\
\bottomrule
\end{tabular}
}
\caption{Statistics of the TFQ-Data and TFQ-Bench splits.}
\vspace{-12pt}
\label{tab:dataset_splits}
\end{table}

\subsection{TFQ-GRPO}

Effectively training models for open-style image implication reasoning presents a significant design challenge. Directly training on OSQ is difficult due to its chaotic, high-dimensional search space and sparse reward signals. While MCQ is more structured, it also suffers from lower knowledge density and sparse rewards, making learning inefficient. We posit that our TFQ format is an ideal training mechanism for this task. The TFQ offers high knowledge density, a graduated difficulty spectrum (from easy to hard), and easily verifiable answers, providing a dense and stable learning signal for reinforcement learning.

We therefore propose TFQ-GRPO, a framework that leverages the TFQ-Data to fine-tune the model's reasoning capabilities. For the optimization algorithm, we adopt Group Relative Policy Optimization (GRPO), which has proven effective for diverse tasks. 

\textbf{Reward Design.}
In multimodal environments, sparse and outcome-driven reward signals are crucial for guiding vision-language models toward effective reasoning and decision-making. Given the open-style thinking process of the image implication question, we adopt a reward formulation that evaluates the reasoning trajectory based on final outcome quality and thinking format.
The total reward is composed of two parts: the accuracy reward $R_{acc}$ and the formatting reward $R_{format}$. The accuracy reward assesses whether the final answer is correct, while the format reward penalizes poorly structured outputs. Formally, given a reasoning trajectory $\tau$, the total reward is defined as:
{\small
\begin{equation}
\label{eq:reward}
R(\tau) = \alpha R_{acc}(\tau) + (1-\alpha) R_{format}(\tau)
\end{equation}
}
where $R_{acc}$ is a binary reward for the correct final answer, $R_{format}$ is a penalty for outputs that do not adhere to the specified tag structure, and $\alpha \in [0, 1]$ is a hyperparameter balancing their importance.

\textbf{GRPO.} GRPO is an on-policy reinforcement learning algorithm. For each input $x$, the old policy model $\pi_{\theta_{old}}$ from previous step generate a group of rollouts $\{o_{i}\}^G_{i=1}$. Then, our reward function is used to calculate rewards for each $o_i$, getting $\{r_{i}\}^G_{i=1}$. We design a unified reward mechanism and the relative advantage is calculated as:
\begin{equation}
\footnotesize
    A_i = \frac{r_i - \text{mean}(\{r_1, r_2, \dots, r_G\})}{\text{std}(\{r_1, r_2, \dots, r_G\})}.
\end{equation}
GRPO maximizes the following objective to optimize the model $\pi_\theta$:
{\small
\begin{equation}
\label{eq:objective}
\footnotesize
\begin{split}
&\mathcal{J}_{\text{GRPO}}(\theta) = \mathbb{E}_{x \sim \text{Train Batch},\, \{o_i\}_{i=1}^G \sim \pi_{\theta_{\text{old}}}(O|x)} \\
&\quad \left[ \frac{1}{G} \sum_{i=1}^G \min\left( \frac{\pi_\theta(o_i\mid x)}{\pi_{\theta_{\text{old}}}(o_i\mid x)} A_i,\, \text{clip}\left( \frac{\pi_\theta(o_i\mid x)}{\pi_{\theta_{\text{old}}}(o_i\mid x)}, 1-\varepsilon, 1+\varepsilon \right) A_i \right) \right. \\
&\quad \left. - \beta D_{\text{KL}}(\pi_\theta \parallel \pi_{\text{ref}}) \vphantom{\sum_{i=1}^G} \right].
\end{split}
\end{equation}
}

The core component of TFQ-GRPO is the structured reasoning prompt that guides the model through the desired inferential logic: \textit{Image Description} $\rightarrow$ \textit{Implication Analysis} $\rightarrow$ \textit{Final Answer}. We instruct the model to first describe the image, then analyze its implications, and finally reason to get the answer. Our training template is shown in Table~\ref{tab:prompt_tfq-grpo}. 

\begin{table}[hbtp]
\centering
\begin{tabular}{p{0.9\linewidth}} % 单列，宽度适配双栏
\toprule
\textbf{SYSTEM:} Please according to the image, and try to answer the following true-false questions with the option T (True) or F (False). \textcolor{darkred}{First, describe the image, then analyze the image implication, and finally reason to get the answer}. Output the thinking process in \verb|<think>|\verb|</think>| and the final correct answer in \verb|<answer>|\verb|</answer>| tags. The output format should be as follows: \verb|<think>|...\verb|</think>| \verb|<answer>|...\verb|</answer>|. \\
\textbf{USER:} True-false questions: \verb|{}| \\
\bottomrule
\end{tabular}
\caption{Training Template of TFQ-GRPO.}
\vspace{-12pt}
\label{tab:prompt_tfq-grpo}
\end{table}

\section{MetaphorStar Family}
\label{sec:metaphorstar}
We introduce the MetaphorStar family, which comprises three sizes: 3B, 7B, and 32B. We utilize the QwenVL-2.5 series as the base model. We provide a detailed analysis of these models in the following sections.

\subsection{Training Setup}

We train all MetaphorStar models using an end-to-end TFQ-GRPO. We initially investigate a conventional two-stage pipeline, which involves a Supervised Fine-Tuning (SFT) warmup stage before RL. However, we find this method suboptimal, as the SFT warmup tends to constrain the model's intrinsic reasoning capabilities. The details are in Sec.~\ref{sec:ablation} and Sec.~\ref{sec:discussion}.
In contrast, training directly with end-to-end TFQ-GRPO yields superior performance and better generalization. Therefore, we adopt the direct end-to-end RL for all experiments. The training process leverages the TFQ-Data-Lite.
For the TFQ-GRPO algorithm, we set the group size for rollouts to $G=5$. In our reward formulation, the hyperparameter $\alpha$ that balances the accuracy ($R_{acc}$) and format reward ($R_{format}$) is set to $0.5$.

\vspace{-3pt} 
\subsection{Analyzing Token Entropy in Reasoning}
\begin{figure}[tp]
    \centering
	\includegraphics[width=0.95\linewidth]{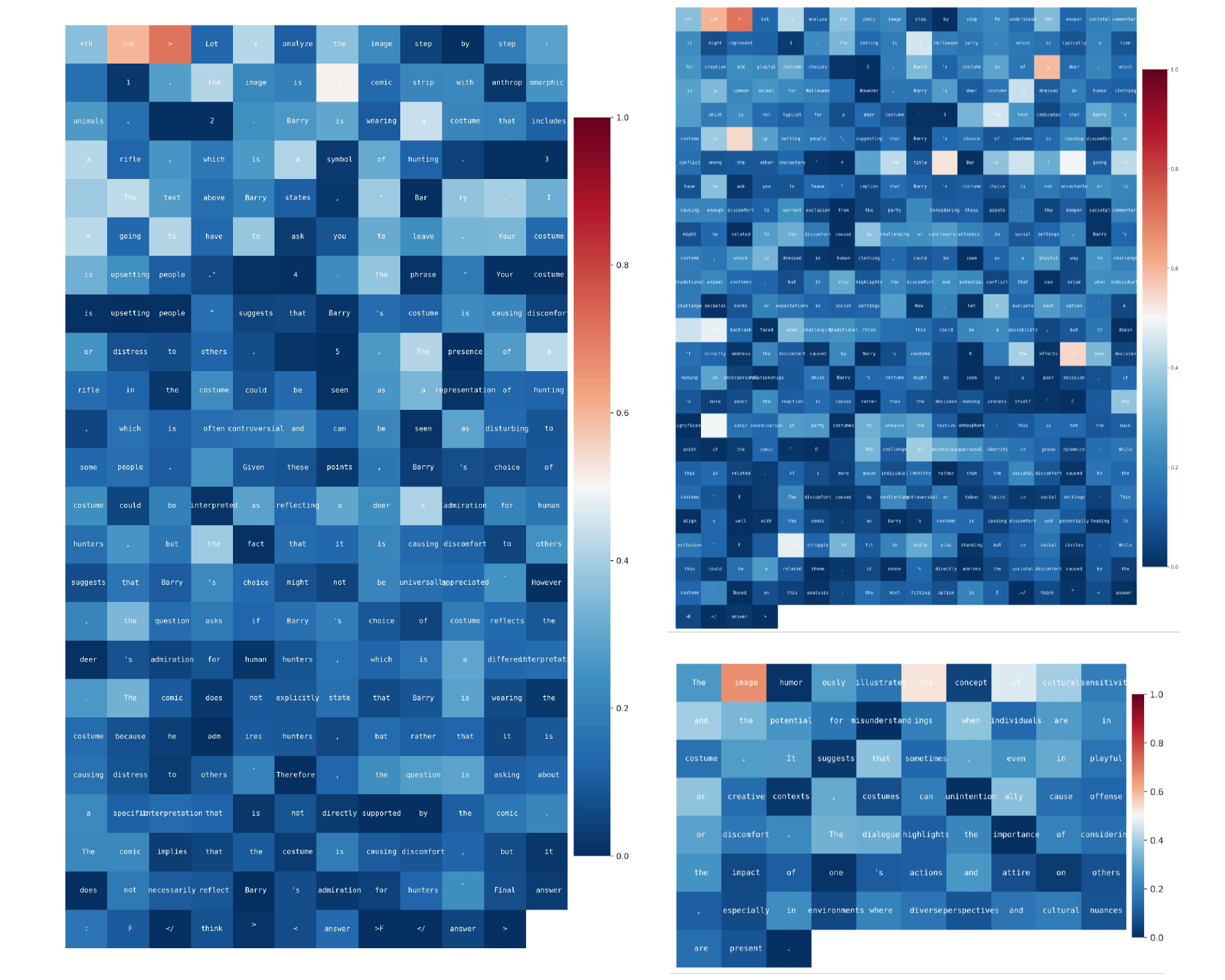}
    \vspace{-5pt}
    \caption{The visualization of token entropy for MetaphorStar-7B on TFQ, MCQ, and OSQ. High-entropy (red) indicates high uncertainty, while low-entropy (blue) indicates high confidence.}
    \label{fig:metaphorstar_heatmap}
    \vspace{-10pt}    
\end{figure}

% \vspace{-10pt}  
To gain insight into the internal reasoning mechanisms of our model, we analyze its token-level generation entropy. Figure~\ref{fig:metaphorstar_heatmap} provides a visualization of this entropy as MetaphorStar-7B generates responses for the TFQ, MCQ, and OSQ tasks.
Our analysis reveals that high-entropy tokens, representing points of highest uncertainty for the model, are not randomly distributed. This aligns with recent findings that ``high-entropy minority" of tokens is critical for complex reasoning \citep{wang20258020rulehighentropyminority}. In the context of image implication, we observe that these spikes in uncertainty consistently occur at crucial semantic and logical junctions.

Specifically, the model exhibits high entropy when generating logical connectors (e.g., ``therefore", ``thus", ``but") that pivot the argument or establish a causal link. We also note high entropy for key function words (e.g., ``the", ``is"), quantifiers, and pronouns, suggesting that the model's core cognitive effort is concentrated on making definitive logical leaps and structuring the relationship between concepts. Conversely, low-entropy (high-confidence) tokens are typically associated with reproducing factual details from the image or completing deterministic phrasal structures.

\section{Experiment}
\label{sec:experiments}
% 主实验 - TFQ + MCQ + OSQ
\begin{table*}[!th]
\centering
\scalebox{0.8}{
\setlength{\tabcolsep}{10pt} % Adjusted spacing
\begin{tabular}{lc|cc} 
\toprule
% --- Header Row 1: Main Categories ---
\textbf{Model} & \textbf{True-False Question} & \textbf{Multiple-Choice Question} & \textbf{Open-Style Question}\\ 
% --- Header Row 2: Sub-categories (en/zh) with rules underneath ---
\midrule
\multicolumn{4}{c}{\textit{General Models}} \\
\midrule
QwenVL-2.5-3B \citep{qwen25vl} & 20\% & 48\% & 2.44 \\  
LLaVA-1.5-7B \citep{liu2023improvedllava} & 0\% & 16\% & 2.06 \\
QwenVL-2.5-7B \citep{qwen25vl} & 28\% & 46\% & 2.34 \\  
DeepSeek-VL2 \citep{wu2024deepseekvl2mixtureofexpertsvisionlanguagemodels} & 20\% & 46\% & 2.82 \\ 
GLM-4.1V-8B \citep{glm4v} & 38\% & 60\% & 2.60 \\  
QwenVL-2.5-32B \citep{qwen25vl} & 56\% & 62\% & 3.08 \\  
GPT-4o-mini \citep{gpt4o} & 36\% & 44\% & 2.98 \\ 
% Gemini-2.0-flash \citep{gemini2} & 28\% & 70\% & 1.60 \\ 
Gemini-2.5-flash \citep{gemini2} & 56\% & 76\% & 3.34 \\ 
% QwenVL-2.0-72B \citep{qwen2VL} & 44\% & 68\% & 2.84 \\
QwenVL-2.5-72B \citep{qwen25vl} & 50\% & 72\% & 1.56 \\
InternVL3-78B \citep{internvl3} & 36\% & 70\% & 3.42 \\
GLM-4V-plus \citep{glm4v} & 42\% & 64\% & 3.01 \\  
% Gemini-2.0-pro \citep{gemini2} & {x\%} & 68\% & 1.66 \\  
Grok-3 \citep{grok3} & 36\% & 66\% & 3.24 \\ 
Claude-3.5-Sonnet \citep{claude35sonnet} & 38\% & 68\% & 3.22 \\
Claude-4.0-Sonnet \citep{claude4sonnet} & 52\% & 60\% & 3.46 \\
GPT-4o \citep{gpt4o} & 50\% & 74\% & 2.94 \\ 
GPT-4.1 \citep{gpt4.1} & 40\%& 74\% & 3.30 \\  
% GPT-5.1 & X & 80\% & 3.64 \\
\midrule
\multicolumn{4}{c}{\textit{Vision-language Reasoning Models}} \\
\midrule
% Gemini-2.0-flash-thinking \citep{gemini2think} & 28\% & 64\% & 1.66 \\
Gemini-2.5-flash-thinking \citep{gemini2think} & 54\% & 78\% & 3.42 \\  
QVQ-72B \citep{qvq-72b-preview} & 28\% & 62\% & 3.10 \\
o4-mini \citep{o4-mini} & 42\% & 58\% & 3.26 \\
Doubao-1.5-thinking-vision-pro \citep{doubao-1.5-thinking-vision-pro} & 62\% & 66\% & 3.16 \\
Grok-3-reasoning \citep{grok3} & 36\% & 74\% & 3.06 \\
% Gemini-2.5-pro \citep{gemini2} & 68\% & \textbf{82\%} & 3.38 \\  
Gemini-3.0-pro \citep{gemini3} & 58\% & \underline{76\%} & \underline{3.82} \\  
\midrule
\multicolumn{4}{c}{\textit{Our MetaphorStar Family}} \\
\midrule
MetaphorStar-3B & 62\% & 64\% & 3.06 \\
MetaphorStar-7B & \underline{70\%} & 74\% & 3.22 \\
MetaphorStar-32B & \textbf{74\%} & \textbf{78\%} & \textbf{3.94} \\
% --- Added Improvement Row ---
% Improv. & \textcolor{darkdarkred}{+42 (150\%)} & \textcolor{darkdarkred}{+28 (60.1\%)} & \textcolor{darkdarkred}{+0.88 (37.6\%)} \\  % Added calculated improvement. Let's show both absolute (pp) and relative (%) for clarity.
\bottomrule
\end{tabular}
} 
\caption{Overall results of different models on True-False Question, Multiple-Choice Question and Open-Style Question. The best-performing model in each category is \textbf{in-bold}, and the second best is \underline{underlined}.}
\vspace{-12pt}
\label{tab:main_results} 
\end{table*}

% generalization experiment
\begin{table*}[!t]
\centering
\scalebox{0.7}{
\setlength{\tabcolsep}{10pt} % Adjusted spacing
\begin{tabular}{l|lll|ccc} 
\toprule
% --- Header Row 1: Main Categories ---
\multirow{2}{*}{\textbf{Benchmark}} & \multicolumn{3}{c|}{\textbf{MetaphorStar Family}} &  \multicolumn{3}{c}{\textbf{Base Model}} \\
\cmidrule(lr){2-4} \cmidrule(lr){5-7}
& \textbf{MetaphorStar-32B} & \textbf{MetaphorStar-7B} & \textbf{MetaphorStar-3B} & \textbf{QwenVL-2.5-32B} & \textbf{QwenVL-2.5-7B} & \textbf{QwenVL-2.5-3B} \\ 
% --- Header Row 2: Sub-categories (en/zh) with rules underneath ---
\midrule
\multicolumn{7}{c}{\textit{Reasoning}} \\
\midrule
MMMU$_{\text{test}}$ & \textbf{49.8}$_{\textcolor{darkgreen}{\uparrow\textcolor{darkgreen}{16.2}}}$ & 48.8$_{\textcolor{darkgreen}{\uparrow\textcolor{darkgreen}{6.8}}}$ & 45.1$_{\textcolor{darkgreen}{\uparrow\textcolor{darkgreen}{2.8}}}$ & 33.6 & 42.0 & 42.3 \\  
VisualPuzzles & \textbf{39.7}$_{\textcolor{darkgreen}{\uparrow\textcolor{darkgreen}{2.5}}}$ & 35.9$_{\textcolor{darkgreen}{\uparrow\textcolor{darkgreen}{2.2}}}$ & 33.8$_{\textcolor{darkgreen}{\uparrow\textcolor{darkgreen}{2.8}}}$ & 37.2 & 33.7 & 31.0 \\
LogicVista & \textbf{56.6}$_{\textcolor{darkgreen}{\uparrow\textcolor{darkgreen}{1.6}}}$ & 47.2$_{\textcolor{darkgreen}{\uparrow\textcolor{darkgreen}{3.1}}}$ & 39.4 & 55.0 & 44.1 & 39.4 \\
VisuLogic & 25.5$_{\textcolor{darkred}{\downarrow\textcolor{darkred}{0.8}}}$ & \textbf{26.9}$_{\textcolor{darkgreen}{\uparrow\textcolor{darkgreen}{2.2}}}$ & 18.8$_{\textcolor{darkred}{\downarrow\textcolor{darkred}{0.3}}}$ & 26.3 & 24.7 & 19.1 \\
V* & \textbf{81.2}$_{\textcolor{darkgreen}{\uparrow\textcolor{darkgreen}{0.1}}}$ & 76.4$_{\textcolor{darkgreen}{\uparrow\textcolor{darkgreen}{5.2}}}$ & 34.0 & 81.1 & 71.2 & 34.0 \\
ZeroBench$_{\text{main}}$ & \textbf{1.0}$_{\textcolor{darkgreen}{\uparrow\textcolor{darkgreen}{1.0}}}$ & \textbf{1.0}$_{\textcolor{darkgreen}{\uparrow\textcolor{darkgreen}{1.0}}}$ & 0.0 & 0.0 & 0.0 & 0.0 \\
ZeroBench$_{\text{sub}}$ & \textbf{18.0}$_{\textcolor{darkgreen}{\uparrow\textcolor{darkgreen}{2.4}}}$ & 15.3$_{\textcolor{darkgreen}{\uparrow\textcolor{darkgreen}{1.2}}}$ & 6.6$_{\textcolor{darkgreen}{\uparrow\textcolor{darkgreen}{1.2}}}$ & 15.6 & 14.1 & 5.4 \\
MathVision & \textbf{38.1}$_{\textcolor{darkgreen}{\uparrow\textcolor{darkgreen}{0.7}}}$ & 25.3$_{\textcolor{darkgreen}{\uparrow\textcolor{darkgreen}{0.2}}}$ & 22.2$_{\textcolor{darkgreen}{\uparrow\textcolor{darkgreen}{1.0}}}$ & 37.4 & 25.1 & 21.2 \\
MathVerse$_{\text{Vision}}$ & \textbf{50.8}$_{\textcolor{darkgreen}{\uparrow\textcolor{darkgreen}{2.4}}}$ & 41.4$_{\textcolor{darkgreen}{\uparrow\textcolor{darkgreen}{6.2}}}$ & 30.0$_{\textcolor{darkgreen}{\uparrow\textcolor{darkgreen}{0.8}}}$ & 48.4 & 35.2 & 29.2 \\  
WeMath & \textbf{48.6}$_{\textcolor{darkgreen}{\uparrow\textcolor{darkgreen}{2.5}}}$ & 36.7$_{\textcolor{darkgreen}{\uparrow\textcolor{darkgreen}{2.4}}}$ & 21.7$_{\textcolor{darkred}{\downarrow\textcolor{darkred}{1.2}}}$ & 46.1 & 34.3 & 22.9 \\
\rowcolor{lightblue!30}
Avg. & \textbf{41.0}$_{\textcolor{darkgreen}{\uparrow\textcolor{darkgreen}{2.9}}}$ & 35.5$_{\textcolor{darkgreen}{\uparrow\textcolor{darkgreen}{3.2}}}$ & 25.4$_{\textcolor{darkgreen}{\uparrow\textcolor{darkgreen}{1.0}}}$ & 38.1 & 32.3 & 24.4 \\
\midrule
\multicolumn{7}{c}{\textit{Understanding}} \\
\midrule
SEEDBench & \textbf{77.6}$_{\textcolor{darkgreen}{\uparrow\textcolor{darkgreen}{0.2}}}$ & 77.1$_{\textcolor{darkgreen}{\uparrow\textcolor{darkgreen}{0.1}}}$ & 74.0 & 77.4 & 77.0 & 74.0 \\  
SEEDBench2 Plus & \textbf{73.2}$_{\textcolor{darkgreen}{\uparrow\textcolor{darkgreen}{0.8}}}$ & 70.8$_{\textcolor{darkgreen}{\uparrow\textcolor{darkgreen}{0.1}}}$ & 63.6$_{\textcolor{darkgreen}{\uparrow\textcolor{darkgreen}{0.3}}}$ & 72.4 & 70.7 & 63.3 \\  
MMBench-V1.0-EN$_{\text{test}}$ & 85.8$_{\textcolor{darkred}{\downarrow\textcolor{darkred}{0.6}}}$ & 83.5 & 79.7$_{\textcolor{darkgreen}{\uparrow\textcolor{darkgreen}{0.6}}}$ & \textbf{86.4} & 83.5 & 79.1 \\
MMBench-V1.1-EN$_{\text{test}}$ & \textbf{84.4}$_{\textcolor{darkgreen}{\uparrow\textcolor{darkgreen}{0.4}}}$ & 82.5$_{\textcolor{darkgreen}{\uparrow\textcolor{darkgreen}{0.3}}}$ & 77.6$_{\textcolor{darkgreen}{\uparrow\textcolor{darkgreen}{0.8}}}$ & 84.0 & 82.2 & 76.8 \\ 
MMStar & \textbf{68.5}$_{\textcolor{darkgreen}{\uparrow\textcolor{darkgreen}{2.2}}}$ & 64.1$_{\textcolor{darkgreen}{\uparrow\textcolor{darkgreen}{0.2}}}$ & 55.5$_{\textcolor{darkred}{\downarrow\textcolor{darkred}{0.4}}}$ & 66.3 & 63.9 & 55.9 \\  
OCRBench & 86.1$_{\textcolor{darkgreen}{\uparrow\textcolor{darkgreen}{0.5}}}$ & \textbf{88.6}$_{\textcolor{darkgreen}{\uparrow\textcolor{darkgreen}{2.2}}}$ & 81.8$_{\textcolor{darkgreen}{\uparrow\textcolor{darkgreen}{2.1}}}$ & 85.6 & 86.4 & 79.7 \\ 
AI2D$_{\text{test}}$ & 83.3$_{\textcolor{darkgreen}{\uparrow\textcolor{darkgreen}{1.0}}}$ & \textbf{84.4}$_{\textcolor{darkgreen}{\uparrow\textcolor{darkgreen}{0.5}}}$ & 81.2$_{\textcolor{darkgreen}{\uparrow\textcolor{darkgreen}{0.4}}}$ & 82.3 & 83.9 & 80.8 \\
ScienceQA$_{\text{test}}$ & \textbf{91.3}$_{\textcolor{darkgreen}{\uparrow\textcolor{darkgreen}{0.4}}}$ & 89.0 & 81.8$_{\textcolor{darkgreen}{\uparrow\textcolor{darkgreen}{0.4}}}$ & 90.9 & 89.0 & 81.4 \\  
POPE & \textbf{86.3}$_{\textcolor{darkgreen}{\uparrow\textcolor{darkgreen}{0.6}}}$ & 86.0$_{\textcolor{darkgreen}{\uparrow\textcolor{darkgreen}{0.1}}}$ & 86.2$_{\textcolor{darkgreen}{\uparrow\textcolor{darkgreen}{0.3}}}$ & 85.7 & 85.9 & 85.9 \\  
MMT-Bench$_{\text{val}}$ & \textbf{65.7}$_{\textcolor{darkgreen}{\uparrow\textcolor{darkgreen}{0.4}}}$ & 62.3$_{\textcolor{darkgreen}{\uparrow\textcolor{darkgreen}{0.2}}}$ & 56.6$_{\textcolor{darkgreen}{\uparrow\textcolor{darkgreen}{0.3}}}$ & 65.3 & 62.1 & 56.3 \\ 
RealworldQA$_{\text{avg}}$ & \textbf{71.1}$_{\textcolor{darkgreen}{\uparrow\textcolor{darkgreen}{1.0}}}$ & 68.1$_{\textcolor{darkred}{\downarrow\textcolor{darkred}{0.4}}}$ & 62.4$_{\textcolor{darkred}{\downarrow\textcolor{darkred}{3.0}}}$ & 70.1 & 68.5 & 65.4 \\
BLINK$_{\text{val}}$ & 62.9$_{\textcolor{darkred}{\downarrow\textcolor{darkred}{0.8}}}$ & 56.6$_{\textcolor{darkgreen}{\uparrow\textcolor{darkgreen}{1.3}}}$ & 44.5$_{\textcolor{darkred}{\downarrow\textcolor{darkred}{0.2}}}$ & \textbf{63.7} & 55.3 & 44.7 \\
HallusionBench$_{\text{avg}}$ & 55.3$_{\textcolor{darkred}{\downarrow\textcolor{darkred}{1.4}}}$ & 49.9$_{\textcolor{darkred}{\downarrow\textcolor{darkred}{1.8}}}$ & 46.5$_{\textcolor{darkgreen}{\uparrow\textcolor{darkgreen}{0.2}}}$ & \textbf{56.7} & 51.7 & 46.3 \\
MMVet Hard & 59.6$_{\textcolor{darkred}{\downarrow\textcolor{darkred}{4.4}}}$ & 54.6$_{\textcolor{darkgreen}{\uparrow\textcolor{darkgreen}{1.7}}}$ & 50.2$_{\textcolor{darkred}{\downarrow\textcolor{darkred}{0.5}}}$ & \textbf{64.0} & 52.9 & 50.7 \\
\rowcolor{lightblue!30}
Avg. & \textbf{75.1}$_{\textcolor{darkgreen}{\uparrow\textcolor{darkgreen}{0.1}}}$ & 72.7$_{\textcolor{darkgreen}{\uparrow\textcolor{darkgreen}{0.3}}}$ & 67.3$_{\textcolor{darkgreen}{\uparrow\textcolor{darkgreen}{0.1}}}$ & 75.0 & 72.4 & 67.2 \\
\midrule
Overall Avg. & \textbf{60.9}$_{\textcolor{darkgreen}{\uparrow\textcolor{darkgreen}{1.3}}}$ & 57.2$_{\textcolor{darkgreen}{\uparrow\textcolor{darkgreen}{1.5}}}$ & 49.9$_{\textcolor{darkgreen}{\uparrow\textcolor{darkgreen}{0.5}}}$ & 59.6 & 55.7 & 49.4 \\
\bottomrule
\end{tabular}
}
\caption{Results on different visual question answering tasks. The best-performing model in each category is \textbf{in-bold}. Performance differences relative to base models are shown as colored subscripts: $\textcolor{darkgreen}{_{\uparrow}}$ for improvements, $\textcolor{darkred}{_{\downarrow}}$ for declines.}
\vspace{-12pt}
\label{tab:other_results} 
\end{table*}

% 主实验
\subsection{Main Experiment}

% \subsubsection{Implementation Details}

% \textbf{Models and Evaluation.}
We carefully select a diverse range of MLLMs.
Our evaluation utilizes the TFQ-Bench-Lite. 
For the comprehensive evaluation on image implication tasks, we also test on the high-level bench (EN) \citep{zhang2025letandroidsdream}, featuring Multiple-Choice Question (MCQ) and Open-Style Question (OSQ). The details are in Appendix~\ref{app:experiment_details}.

\subsubsection{True-False Question}
% 1.MetaphorStar-32B正确率是所有模型第一（正确率74%），同时MetaphorStar-7B正确率第二（正确率70%），都超过了top闭源模型Gemini-2.5-pro；令人惊讶的是，MetaphorStar-3B超过了Claude-4.0-Sonnet和GPT-4.1得分较多，表明现有顶级模型在图片隐喻理解任务上的严重不足；
% 2.MetaphorStar-7B相对基模型QwenVL-2.5-7B提升了150%，MetaphorStar-3B相对基模型QwenVL-2.5-3B提升了210%，证明了我们的数据集TFQ-Data和训练方法TFQ-GRPO的强有效性；
% 3.我们发现，在TFQ任务上，reasoning models比general models普遍较强。我们推测是因为TFQ任务包含部分basic VQA tasks which robe the model’s grasp of primary visual information，xxx.
Table~\ref{tab:main_results} presents comprehensive results of TFQ across different MLLMs on the TFQ-Bench-Lite. Our MetaphorStar family achieves SOTA performance. Surprisingly, MetaphorStar-3B (62\%) also surpasses the powerful closed-source model Gemini-3.0-pro (58\%), indicating a severe deficiency in existing top-tier MLLMs for this task.

The effectiveness of our training is stark. MetaphorStar-7B (70\%) shows a 150\% relative improvement over its QwenVL-2.5-7B base (28\%), and MetaphorStar-3B (62\%) achieves a 210\% relative gain over its QwenVL-2.5-3B base (20\%). This demonstrates the potent efficacy of our TFQ-Data and the TFQ-GRPO method. We also observe that reasoning models generally perform better than general models on TFQ, which we attribute to the task's inclusion of basic VQA-style questions that probe primary visual information.

\subsubsection{Multiple-Choice Question}

Table~\ref{tab:main_results} presents comprehensive results of MCQ across different MLLMs on the high-level bench (EN). Our models demonstrate strong generalization.
MetaphorStar-32B is the top-performing open-source model, outperforming closed-source top models Gemini-3.0-pro (76\%). 
The generalization from our TFQ-centric training is evident. MetaphorStar-7B (74\%) achieves a 60\% relative improvement over its base model (46\%), and MetaphorStar-3B (64\%) achieves a 34\% relative improvement over its base model (48\%).
Notably, on this task, the distinction between ``reasoning" and "general" models is minimal.
This suggests that the RL-based training in many existing reasoning models (often focused on math or code) has limited generalization to the abstract domain of image implication, which again highlights the unique effectiveness of our TFQ-GRPO.

\subsubsection{Open-Style Question}

Table~\ref{tab:main_results} presents comprehensive results of OSQ across different MLLMs on the high-level bench (EN).
On the highly challenging OSQ task, MetaphorStar-32B (3.94) achieves the best score, significantly outperforming all other models, including Gemini-3.0-pro (3.82), Claude-4.0-Sonnet (3.46).
This further proves the robust generalization of our method. MetaphorStar-7B shows a 38\% relative gain over its base.
Interestingly, unlike the MCQ results, we see significant performance disparities between reasoning and general models on OSQ.
We also note that some models (e.g., QwenVL-2.5-72B) perform well on MCQ but poorly on OSQ.
We attribute this to potential overfitting to multiple-choice formats and insufficient exposure to open-style generation. 
In addition, LLMs or even MLLMs may not genuinely understand the questions but rather predict options as answers, having evaluation bias and demonstrating sensitivity to option positioning, with similar findings in \citep{zhang2025letandroidsdream}.

\subsection{Generalization Experiment}

\subsubsection{Benchmarks and baselines}
% Understanding and Reasoning 
% We mainly select two categories of benchmarks — Reasoning and Understanding. Reasoning ability is quite critical for MLLMs' deployment in environments requiring complex decision-making. Understanding ability is crucial for MLLMs' robustness and generalization in the real world.
% The detailed benchmarks are listed in Appendix \ref{app:experiment_details}. 
% To provide a more intuitive view of the model's overall performance across domains, we compute the average score, normalized from 0 to 100, where higher is better.
% We take QwenVL-2.5 series~\citep{qwen25vl} as the primary baseline. Our models include the MetaphorStar family. To ensure a fair comparison, we employ VLMEvalKit~\citep{duan2024vlmevalkit} as the evaluation pipeline.

We evaluate generalization across two benchmark categories: (1) {Reasoning}, which is critical for complex decision-making, and (2) {Understanding}, which is crucial for real-world robustness. 
Appendix \ref{app:experiment_details} lists the specific benchmarks. 
For a high-level overview, we report an average score (normalized 0--100, higher is better) across all benchmarks. We compare our MetaphorStar models against the QwenVL-2.5 series~\citep{qwen25vl} baselines. To ensure a fair comparison, all evaluations employ VLMEvalKit~\citep{duan2024vlmevalkit}.

\subsubsection{Evaluation Results}

Table~\ref{tab:other_results} details the generalization performance of the MetaphorStar family against their respective base models. The results verify that our training on the image implication task provides a significant boost to visual reasoning, while simultaneously maintaining or even slightly improving performance on general visual understanding tasks, demonstrating robust and targeted generalization. Please see the detailed analysis in Sec. \ref{sec:discussion}.

\textbf{Reasoning.}
The MetaphorStar family shows substantial and consistent reasoning improvements. On average, MetaphorStar-7B improves by 3.2 points and MetaphorStar-32B by 2.9 points over their baselines. The gains are most pronounced on challenging benchmarks: MetaphorStar-32B achieves a +16.2 point increase on MMMU (with 7B at +6.8 and 3B at +2.8). Strong gains also appear on MathVerse (+6.2 for 7B) and V* (+5.2 for 7B). This suggests our task's complex, multi-hop inference enhances underlying logical and visual reasoning faculties.

\textbf{Understanding.} 
In this domain, our specialized training does not harm and often slightly improves general visual understanding. The MetaphorStar family maintains stable performance, with slight average improvements (e.g., +0.3 points for MetaphorStar-7B) across the 14 benchmarks. We note positive gains on challenging benchmarks like MMStar (+2.2 for 32B) and OCRBench (+2.2 for 7B). The overall performance confirms our method enhances reasoning without sacrificing foundational understanding.

\section{Ablation Study}
\label{sec:ablation}

\subsection{Model Parameter Scaling}

\begin{figure}[th]
    \centering
	\includegraphics[width=1\linewidth]{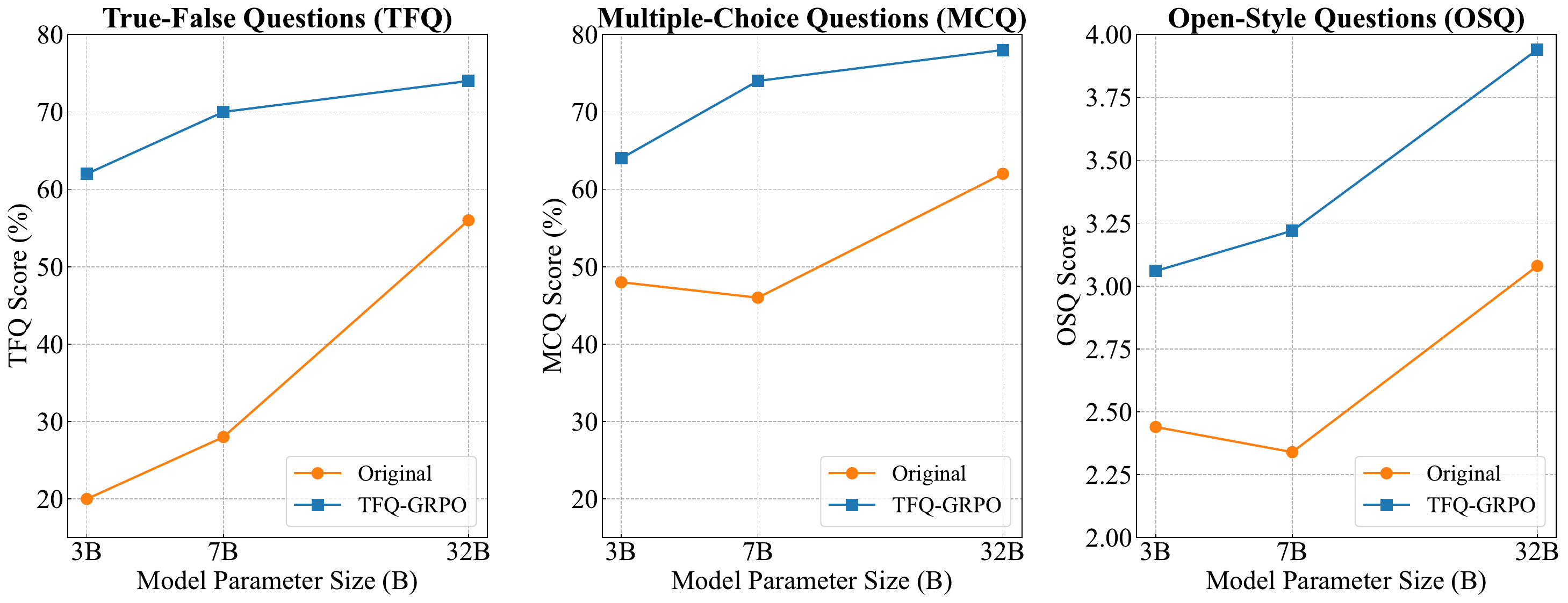}
    \vspace{-10pt}
    \caption{The model parameter scaling law.}
    \vspace{-15pt}
    \label{fig:abl_model_parameter_scaling}
\end{figure}

We analyze the impact of model parameter scaling, with results in Figure~\ref{fig:abl_model_parameter_scaling} and Table~\ref{tab:main_results}. Our analysis reveals that the TFQ-GRPO training is crucial for unlocking the benefits of model scaling. Base models (w/o TFQ-GRPO) exhibit inconsistent or weak scaling; for instance, on OSQ, the 7B base model (2.34) underperforms the 3B base model (2.44). In sharp contrast, our trained MetaphorStar models demonstrate a clean and monotonic performance increase with scale (3.06 $\to$ 3.22 $\to$ 3.94).
With our method enabling effective scaling, we observe that performance systematically improves with parameter count. This effect is most pronounced on OSQ, which shows accelerating marginal returns (3B$\to$7B: +0.16 vs. 7B$\to$32B: +0.72). This suggests that OSQ's open-ended reasoning disproportionately benefits from larger model capacity. Conversely, the closed-ended TFQ and MCQ tasks show more linear, though still consistent, performance gains as model size increases.

% 图表：scaling图

\begin{table*}[!t]
\begin{minipage}[t]{0.25\linewidth}
    \vspace{0pt} % 强制从顶部对齐
    % \centering
    % \small
    % \setlength{\tabcolsep}{4.5pt} % 微调列间距适应单栏
    \scalebox{0.7}{
    \setlength{\tabcolsep}{10pt}
    \begin{tabular}{lc|cc} 
    \toprule
    \textbf{Data} & \textbf{TFQ} & \textbf{MCQ} & \textbf{OSQ} \\ 
    \midrule
    Small (0.1k) & 48\% & 64\% & 3.04 \\
    Lite (1k) & 70\% & \textbf{74\%} & 3.22 \\
    Full (1.4k) & \textbf{84\%} & \textbf{74\%} & \textbf{3.48} \\ 
    \bottomrule
    \end{tabular}
    } 
    % \vspace{7pt}
    \caption{Results of scaling training data. The best-performing model in each category is \textbf{in-bold}.}
    \label{tab:abl_training_data_scaling}
    \end{minipage}
\hfill % 保持两表间距
\begin{minipage}[t]{0.28\linewidth}
    \vspace{0pt} % 强制从顶部对齐的关键技巧
    % \centering
    \scalebox{0.7}{
    \setlength{\tabcolsep}{10pt}
    \begin{tabular}{lc|cc} 
    \toprule
    \textbf{Model} & \textbf{TFQ} & \textbf{MCQ} & \textbf{OSQ} \\ 
    \midrule
    \multicolumn{4}{c}{\textit{LLaVA-1.5-7B}} \\
    \midrule
    w/o TFQ-GRPO & 0\% & 16\% & 2.06 \\ 
    w/ TFQ-GRPO & \textbf{6\%} & \textbf{34\%} & \textbf{2.78}  \\ 
    \bottomrule
    \end{tabular}
    } 
    % \vspace{7pt}
    \caption{Results of different base models. The best-performing model in each category is \textbf{in-bold}.}
    \label{tab:abl_different_baselines}
    \end{minipage}
\hfill % 保持两表间距
\begin{minipage}[t]{0.38\linewidth}
    \vspace{0pt} % 强制从顶部对齐的关键技巧
    % \centering
    \scalebox{0.7}{
    \setlength{\tabcolsep}{10pt}
    \begin{tabular}{lc|cc} 
    \toprule
    \textbf{Model} & \textbf{TFQ} & \textbf{MCQ} & \textbf{OSQ} \\ 
    \midrule
    QwenVL-2.5-7B & 28\% & 46\% & 2.34 \\ 
    { } + TFQ-SFT & 42\% & 28\% & 3.34 \\ 
    { } + TFQ-GRPO & \textbf{70\%} & \textbf{74\%} & 3.22 \\ 
    { } + TFQ-SFT \& TFQ-GRPO & 56\% & 28\% & \textbf{3.66} \\ 
    \bottomrule
    \end{tabular}
    } 
    % \vspace{7pt}
    \caption{Results of different training strategies. The best-performing model in each category is \textbf{in-bold}.}
    \label{tab:abl_different_training_method}
\end{minipage}
\vspace{-10pt} % 调整表格下方间距
\end{table*}

\subsection{Training Data Scaling}

We investigate the impact of training data volume on model performance. We create three high-quality data subsets from TFQ-Data at different scales: Small (0.1k images), Lite (1k images), and Full (1.4k images). We train three distinct MetaphorStar-7B models on these datasets using identical TFQ-GRPO training parameters. 
As shown in Table~\ref{tab:abl_training_data_scaling}, the results demonstrate two trends. First, performance scales positively and significantly with data quantity across all three tasks. Second, all three models substantially outperform the QwenVL-2.5-7B base model, confirming the powerful effect of our TFQ-Data.
It is particularly noteworthy that even the MetaphorStar-7B-Small model, trained on only 0.1k images, improving 48\% on TFQ and 64\% on MCQ than base model. This highlights the high quality and data efficiency of our dataset. Furthermore, MetaphorStar-7B-Full, trained on the complete 1.4k dataset, achieves SOTA performance on the TFQ task at 84\%. This result not only leads the 7B scale but also surpasses the 74\% score of the MetaphorStar-32B model (trained on 1k images), underscoring that for this task, data scale can also be critical.

\subsection{Different Model Architecture}

To validate the generalizability of our TFQ-GRPO training framework, we test its effectiveness on a different model architecture. We select LLaVA-1.5-7B \citep{liu2023improvedllava}, which is based on the Vicuna (LLaMA-based), presenting a distinct architecture from the QwenVL series. We train this model using the same TFQ-GRPO training parameters and dataset TFQ-Data-Lite as our MetaphorStar-7B model and use identical evaluation protocols.
The results are presented in Table~\ref{tab:abl_different_baselines}. The base LLaVA-1.5-7B model struggles significantly with the image implication task, scoring 0\% on TFQ, 16\% on MCQ, and 2.06 on OSQ. After applying TFQ-GRPO, the model's performance improves dramatically across all three tasks: TFQ score increases to 6\%, MCQ score more than doubles to 34\% (+18\%), and the OSQ score rises to 2.78 (+0.72). These substantial gains demonstrate that our training method is not a specialized fit for QwenVL but is a robust framework capable of enhancing the reasoning capabilities of diverse MLLM architectures.

\subsection{Different Training Strategy}

We explore the impact of different training strategies by comparing three distinct strategies:
1) TFQ-SFT: Supervised Fine-Tuning only.
2) TFQ-SFT \& TFQ-GRPO: SFT as the warmup, followed by RL.
3) TFQ-GRPO: End-to-end RL, which is the main strategy used for MetaphorStar.
To facilitate SFT, we create {TFQ-Data-Lite-SFT}, a dataset of 984 reasoning trajectories. We prompt Claude-3.7-thinking to generate a detailed reasoning process (image description to implication analysis to final answer) for each of the 984 questions in TFQ-Data-Lite.

\begin{figure}[ht]
    \centering
    % \vspace{-10pt}
	\includegraphics[width=0.8\linewidth]{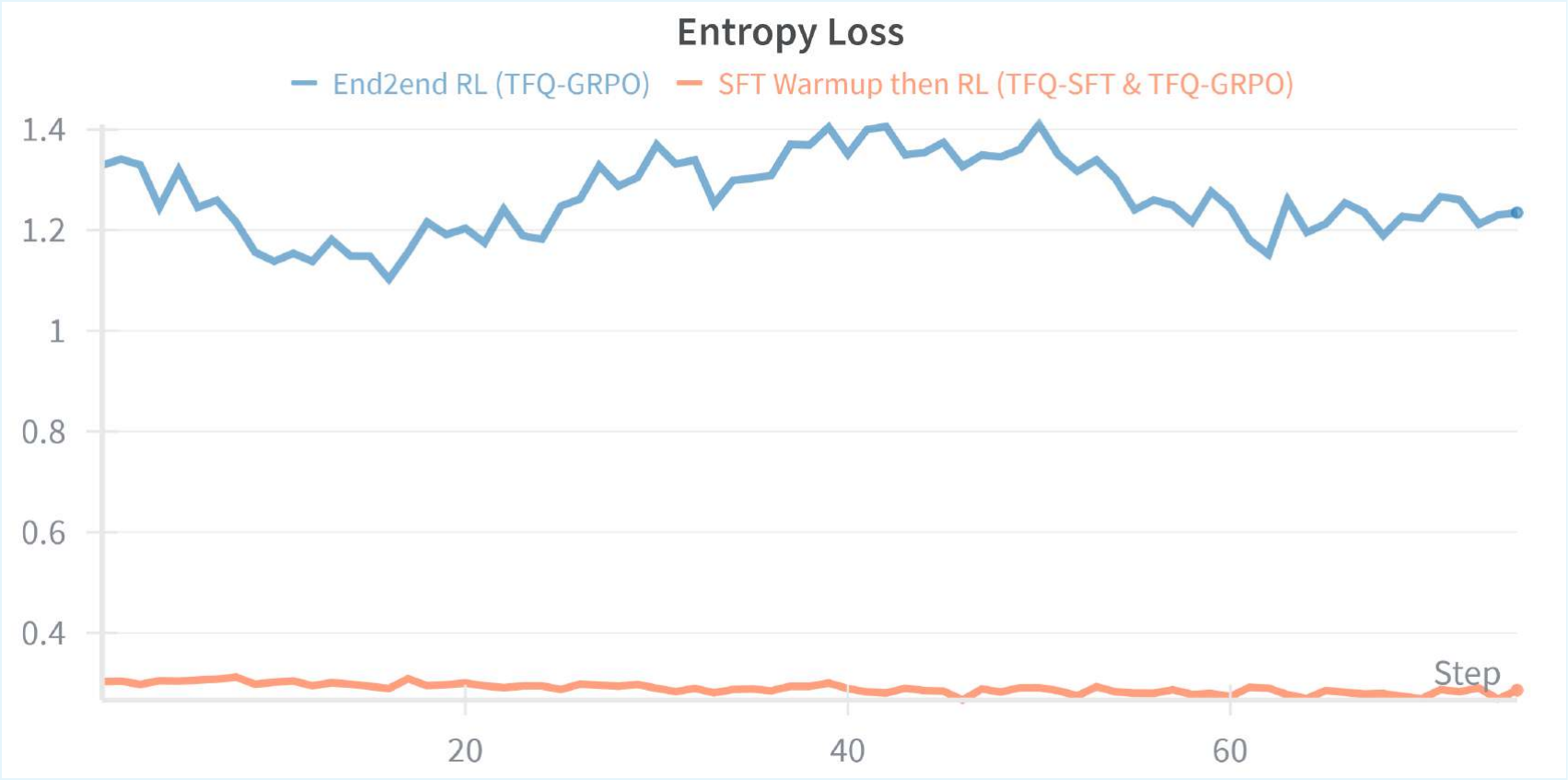}
    \vspace{-5pt}
    \caption{Entropy loss of models with different strategies.}
    \vspace{-10pt}
    \label{fig:abl_model_entropy_loss}
\end{figure}

The results in Table~\ref{tab:abl_different_training_method} lead to a critical finding. Our end-to-end RL method (+ TFQ-GRPO) yields the strongest performance on TFQ and MCQ. Conversely, both strategies involving SFT cause a catastrophic drop in MCQ performance (from 46\% to 28\%), indicating SFT severely damages model generalization.
% This exposes an interesting paradox: the SFT-based methods score highest on the OSQ task (3.66), which is evaluated by the MLLM-as-a-judge. After careful analysis, we find the high score is an artifact. The SFT models learn to be overly verbose, and this verbosity—which often includes multiple, contradictory viewpoints—is misread as comprehensive by the MLLM judge. The end-to-end RL model provides more concise and accurate answers, which receive a lower score from the biased judge.
% Figure~\ref{fig:abl_model_entropy_loss} provides the technical explanation for this phenomenon.
This exposes an important paradox: SFT-based methods score highest on the MLLM-judged OSQ task (3.66). We find the high score is an artifact. SFT models learn to be overly verbose, and this verbosity—which often includes contradictory viewpoints—is misinterpreted as comprehensive by the MLLM judge. Our RL model provides more concise and accurate answers, which are unfairly penalized.
We term this phenomenon the ``SFT Curse", technically explained by token entropy (Figure~\ref{fig:abl_model_entropy_loss}). The base model (1.33) and our end-to-end RL model (1.23) maintain high entropy, allowing for a broad exploration of the solution space. SFT, however, acts as an ``entropy bottleneck," collapsing the model's policy to a low-entropy state (0.30) as it imitates a narrow data distribution. This low-entropy state persists even after RL (0.29), trapping the model in a local optimum focused on stylistic imitation rather than robust reasoning. In contrast, the end-to-end TFQ-GRPO leverages the model's high initial entropy to conduct a broader, more effective search for a global optimum. More details are in Sec.~\ref{sec:discussion}.

% 图表：entropy loss图

\section{Discussion}
\label{sec:discussion}

\subsection{Why SFT Warmup Lose?}
% 具体分析为什么sft warmup 后rl比不上end2end rl的原因或sft warmup是否有必要：1.图片隐喻任务特征  2.TFQ/MCQ/OSQ问题形式 3.entropy角度
% 1.图片隐喻任务特征 
%% 创造性泛化： 实验注意到直接RL模型在图像隐喻任务上“清晰明确，符合预期”，这一点至关重要。抽象和隐喻推理要求模型在不同概念之间建立新颖的联系，这在其预测分布中本质上是低概率事件。SFT通过其最大似然目标，抑制了低概率输出，偏爱来自训练集的高概率“安全”文本。相反，直接RL完全由奖励信号驱动，只要能带来高回报，它就可以自由探索这些创造性路径，从而在非标准的抽象任务上培养出更好的泛化能力。

The ablation in Sec.~\ref{sec:ablation} demonstrates that a conventional SFT warmup stage is not only unnecessary but is actively detrimental to performance on image implication tasks. This phenomenon, which we term the "SFT Curse," stems from a fundamental mismatch between the SFT objective and the nature of the task, as we analyze from three perspectives

\textbf{Task Nature: Creative Generalization.}
Image implication is not a simple pattern recognition task; it demands creative generalization—the ability to connect semantically distant concepts and generate novel, low-probability insights. Supervised Fine-Tuning, as a maximum likelihood objective, directly penalizes this. It trains the model to reproduce the "safe," high-probability sequences from the training data, acting as an "entropy bottleneck" (see Figure~\ref{fig:abl_model_entropy_loss}). This behavioral cloning teaches form over function, trapping the model in a "cognitive straitjacket." In contrast, end-to-end RL is driven purely by the reward signal. It is free to explore and reinforce these creative, low-probability reasoning paths as long as they lead to a correct answer, fostering the robust, abstract reasoning required for metaphors.

\textbf{Question Format: The Talker vs. The Thinker.}
This "form over function" problem is most evident in the MCQ results. TFQ and MCQ are not purely generative tasks; they are highly discriminative. SFT trains the model to be a "talker"—to generate text that sounds plausible and adheres to the structural format (e.g., `<think>...</think>`). It does not, however, train the model to be a "thinker"—to perform the underlying logical discrimination needed to identify and reject incorrect options. This explains the catastrophic collapse in MCQ performance (28\% accuracy) for SFT-warmed models. The end-to-end RL model, by optimizing directly for the accuracy reward ($R_{acc}$), is forced to learn this crucial discriminative capability.

\textbf{The OSQ Paradox: Evaluation Bias.}
This analysis also explains the "OSQ Paradox" in Table~\ref{tab:abl_different_training_method}, where the objectively worse SFT+RL model achieves the highest subjective OSQ score. This is an artifact of the LLM-as-a-judge evaluation. The SFT-trained model produces verbose, well-structured outputs that often mix multiple (and sometimes contradictory) viewpoints. The LLM judge, relying on heuristics, misinterprets this stylistic adherence and verbosity as "deeper thought." The end-to-end RL model, which produces more concise and accurate answers, is unfairly penalized by this bias.

In summary, SFT warmup fails because it creates a low-entropy policy focused on imitation. The subsequent on-policy RL algorithm (GRPO) starts from this skewed distribution and is unable to escape this local optimum. End-to-end RL, by leveraging the high initial entropy of the base model, allows for a true, global search for the optimal reasoning policy.

\subsection{Why Image Implication Tasks Can Help with Visual Reasoning?}

Our generalization experiment (Table~\ref{tab:other_results}) confirms that training on image implication provides significant gains to downstream visual reasoning tasks and even benefits general VQA. We attribute this powerful generalization effect to key properties of the task itself and our training methodology.

\textbf{Cultivating Multi-Hop Abstract Reasoning.}
At its core, image implication is a form of sophisticated, multi-hop abstract reasoning. Unlike standard VQA, which often requires literal, single-hop answers, implication tasks force the model to move from literal perception (e.g., "see a person") to abstract conceptualization (e.g., "understand the person represents a concept") and then to a final conclusion (e.g., "infer the relationship between concepts"). This process of connecting disparate concepts and performing non-literal inference trains the same underlying cognitive faculties required for formal logic, mathematical reasoning, and other complex visual reasoning benchmarks.

\textbf{The Efficacy of the TFQ Format as a Reasoning Trainer.}
The benefits are not just from the what (metaphors) but the how (our TFQ format). As discussed in Section~\ref{sec:method}, TFQ has a high knowledge density, presenting the model with multiple fine-grained propositions to verify for a single image. This transforms the model from a simple "answer generator" into a "propositional verifier." This learned skill of methodically evaluating the truth value of specific claims is a core component that is highly transferable to all logical, mathematical, and sequential reasoning domains.

\textbf{Simultaneous Grounding of Abstraction in Factual Perception.}
Our \texttt{TFQ-Data} design deliberately includes statements that probe basic visual facts alongside the central implication. This dual-objective training ensures that the model does not "drift" into ungrounded abstraction. It learns to simultaneously maintain its core perceptual accuracy (which benefits general VQA) while also building the new scaffolding for abstract inference. This forces the model to learn how to connect concrete visual evidence to abstract logical conclusions, a skill that is central to all robust reasoning.

\section{Conclusion}
\label{sec:conclusion}

% Metaphorical comprehension in images remains a critical challenge for AI systems. We address the critical deficiency of MLLMs in understanding image implication, the task that demands sophisticated, non-literal reasoning. We propose MetaphorStar, a comprehensive visual reinforcement learning framework designed to bridge this gap. Our contribution is threefold: (1) We introduce the True-False Question (TFQ) format, the fine-grained, high-density evaluation approach, and release the corresponding TFQ-Data and TFQ-Bench. (2) We develop the TFQ-GRPO training method, an end-to-end RL strategy that effectively teaches abstract reasoning. (3) We release the MetaphorStar family of models, which achieve SOTA performance on image implication tasks. 
We address the critical challenge of image implication, a form of sophisticated, non-literal reasoning where MLLMs currently struggle. We propose MetaphorStar, the visual reinforcement learning (RL) framework designed to bridge this gap. Our contributions include the True-False Question (TFQ) format for image implication tasks, along with the corresponding TFQ-Data and TFQ-Bench. We also develop TFQ-GRPO, the end-to-end RL training method, and release the MetaphorStar family of models, which achieve SOTA performance. 
Our experiments also reveal two crucial insights. First, image implication tasks can significantly enhance model performance on complex visual reasoning. Second, we identify the ``SFT Curse", demonstrating that traditional SFT warmup creates the "entropy bottleneck" that harms generalization, and show that end-to-end RL methods are more suitable for image implication tasks, even visual reasoning tasks. We open-source all models, datasets, and code to help advance MLLMs beyond literal perception toward deeper, conceptual understanding.

% We address the critical challenge of image implication, a form of sophisticated, non-literal reasoning where MLLMs currently struggle. We propose MetaphorStar, the visual reinforcement learning framework designed to bridge this gap. Our contributions include the True-False Question (TFQ) format for image implication tasks, along with the corresponding TFQ-Data and TFQ-Bench. We also develop TFQ-GRPO, the end-to-end RL training method, and release the MetaphorStar family of models, which achieve SOTA performance on image implication tasks. 

{
    \small
    \bibliographystyle{ieeenat_fullname}
    \bibliography{custom}

@article{dikw,
  title = {Data, Information, Knowledge, Wisdom (DIKW): A Semiotic Theoretical and Empirical Exploration of the Hierarchy and its Quality Dimension},
  author = {Baskarada, Sasa and Koronios, Andy},
  journal = {Australasian Journal of Information Systems},
  year = {2013},
}

@book{lakoff2008metaphors,
  title={Metaphors we live by},
  author={Lakoff, George and Johnson, Mark},
  year={2008},
  publisher={University of Chicago Press}
}

@article{lightman2023letsverifystepstep,
  title={Let's Verify Step by Step}, 
  author={Hunter Lightman and Vineet Kosaraju and Yura Burda and Harri Edwards and Bowen Baker and Teddy Lee and Jan Leike and John Schulman and Ilya Sutskever and Karl Cobbe},
  journal={arXiv preprint arXiv:2305.20050},
  year={2023}
}

@inproceedings{wang2024measuring, 
  title={Measuring Multimodal Mathematical Reasoning with MATH-Vision Dataset},
  author={Ke Wang and Junting Pan and Weikang Shi and Zimu Lu and Mingjie Zhan and Hongsheng Li}, 
  booktitle={NeurIPS}, 
  year={2024}
}

@inproceedings{lu2024mathvista,
  title     = {MathVista: Evaluating Mathematical Reasoning of Foundation Models in Visual Contexts},
  author    = {Lu, Pan and Bansal, Hritik and Xia, Tony and Liu, Jiacheng and Li, Chunyuan and Hajishirzi, Hannaneh and Cheng, Hao and Chang, Kai-Wei and Galley, Michel and Gao, Jianfeng},
  booktitle={ICLR},
  year      = {2024}
}

@inproceedings{yue2023mmmu,
  title={MMMU: A Massive Multi-discipline Multimodal Understanding and Reasoning Benchmark for Expert AGI},
  author={Xiang Yue and Yuansheng Ni and Kai Zhang and Tianyu Zheng and Ruoqi Liu and Ge Zhang and Samuel Stevens and Dongfu Jiang and Weiming Ren and Yuxuan Sun and Cong Wei and Botao Yu and Ruibin Yuan and Renliang Sun and Ming Yin and Boyuan Zheng and Zhenzhu Yang and Yibo Liu and Wenhao Huang and Huan Sun and Yu Su and Wenhu Chen},
  booktitle={CVPR},
  year={2024},
}

@inproceedings{Desai_Chakraborty_Akhtar_2022, 
  title={Nice Perfume. How Long Did You Marinate in It? Multimodal Sarcasm Explanation},
  author={Desai, Poorav and Chakraborty, Tanmoy and Akhtar, Md Shad}, 
  booktitle={AAAI}, 
  year={2022}
}

@inproceedings{hessel-etal-2023-androids,
  title={Do Androids Laugh at Electric Sheep? Humor {``}Understanding{''} Benchmarks from The New Yorker Caption Contest},
  author={Hessel, Jack  and Marasovic, Ana  and Hwang, Jena D.  and Lee, Lillian  and Da, Jeff  and Zellers, Rowan  and Mankoff, Robert and Choi, Yejin},
  booktitle={ACL},
  year={2023}
}

@inproceedings{horvitz-etal-2024-getting,
  title={Getting Serious about Humor: Crafting Humor Datasets with Unfunny Large Language Models},
  author={Horvitz, Zachary and Chen, Jingru and Aditya, Rahul and Srivastava, Harshvardhan and West, Robert and Yu, Zhou and McKeown, Kathleen},
  booktitle={ACL},
  year={2024}
}

@article{zhang2025letandroidsdream,
  title={Let Androids Dream of Electric Sheep: A Human-like Image Implication Understanding and Reasoning Framework}, 
  author={Chenhao Zhang and Yazhe Niu},
  journal={arXiv preprint arXiv:2505.17019},
  year={2025}
}

@inproceedings{yang-etal-2024-large,
  title={Can Large Multimodal Models Uncover Deep Semantics Behind Images?}, 
  author={Yixin Yang and Zheng Li and Qingxiu Dong and Heming Xia and Zhifang Sui},
  booktitle={ACL},
  year={2024}
}

@inproceedings{liu2024iibenchimageimplicationunderstanding,
  title={II-Bench: An Image Implication Understanding Benchmark for Multimodal Large Language Models}, 
  author={Ziqiang Liu and Feiteng Fang and Xi Feng and Xinrun Du and Chenhao Zhang and others},
  booktitle={NeurIPS}, 
  year={2024}
}

@article{zhang2024mllmsunderstanddeepimplication,
  title={Can MLLMs Understand the Deep Implication Behind Chinese Images?}, 
  author={Chenhao Zhang and Xi Feng and Yuelin Bai and Xinrun Du and others},
  journal={arXiv preprint arXiv:2410.13854},
  year={2024}
}

@article{zhong2024lets,
  title={Let's Think Outside the Box: Exploring Leap-of-Thought in Large Language Models with Creative Humor Generation}, 
  author={Shanshan Zhong and Zhongzhan Huang and Shanghua Gao and Wushao Wen and Liang Lin and Marinka Zitnik and Pan Zhou},
  journal={arXiv preprint arXiv:2312.02439},
  year={2024}
}

@inproceedings{xu-etal-2024-exploring,
  title={Exploring Chain-of-Thought for Multi-modal Metaphor Detection}, 
  author={Yanzhi Xu and Yueying Hua and Shichen Li and Zhongqing Wang},
  booktitle={ACL},
  year={2024}
}

@article{li2024vocotunleashingvisuallygrounded,
  title={VoCoT: Unleashing Visually Grounded Multi-Step Reasoning in Large Multi-Modal Models}, 
  author={Zejun Li and Ruipu Luo and Jiwen Zhang and Minghui Qiu and Zhongyu Wei},
  journal={arXiv preprint arXiv:2405.16919},
  year={2024}
}

@article{menon2024whiteboard,
  title={Whiteboard-of-Thought: Thinking Step-by-Step Across Modalities},
  author={Sachit Menon and Richard Zemel and Carl Vondrick},
  journal={arXiv},
  year={2024}
}

@article{hu2024visual,
  title={Visual Sketchpad: Sketching as a Visual Chain of Thought for Multimodal Language Models},
  author={Hu, Yushi and Shi, Weijia and Fu, Xingyu and Roth, Dan and Ostendorf, Mari and Zettlemoyer, Luke and Smith, Noah A and Krishna, Ranjay},
  journal={arXiv preprint arXiv:2406.09403},
  year={2024}
}

@article{xu2024llavacotletvisionlanguage,
  title={LLaVA-CoT: Let Vision Language Models Reason Step-by-Step}, 
  author={Guowei Xu and Peng Jin and Hao Li and Yibing Song and Lichao Sun and Li Yuan},
  journal={arXiv preprint arXiv:2411.10440},
  year={2024}
}

@article{zheng2025deepeyesincentivizingthinkingimages,
  title={DeepEyes: Incentivizing "Thinking with Images" via Reinforcement Learning}, 
  author={Ziwei Zheng and Michael Yang and Jack Hong and Chenxiao Zhao and Guohai Xu and Le Yang and Chao Shen and Xing Yu},
  journal={arXiv preprint arXiv:2505.14362},
  year={2025}
}

@misc{qvq-72b-preview,
  title = {QVQ: To See the World with Wisdom},
  author = {Qwen Team},
  url = {https://qwenlm.github.io/blog/qvq-72b-preview/},
  year = {2024}
}

@misc{o1,
    title = {Learning to reason with llms},
    author = {OpenAI},
    url = {https://openai.com/index/learning-to-reason-with-llms/},
    year = {2024}
}

@article{deepseek-R1,
   title={DeepSeek-R1: Incentivizing Reasoning Capability in LLMs via Reinforcement Learning}, 
   author={DeepSeek-AI and others},
   journal={arXiv preprint arXiv:2501.12948},
   year={2025}
}

@article{yao2024mulberryempoweringmllmo1like,
  title={Mulberry: Empowering MLLM with o1-like Reasoning and Reflection via Collective Monte Carlo Tree Search}, 
  author={Huanjin Yao and Jiaxing Huang and Wenhao Wu and Jingyi Zhang and Yibo Wang and Shunyu Liu and Yingjie Wang and Yuxin Song and Haocheng Feng and Li Shen and Dacheng Tao},
  journal={arXiv preprint arXiv:2412.18319},
  year={2024}
}

@article{wu2024deepseekvl2mixtureofexpertsvisionlanguagemodels,
  title={DeepSeek-VL2: Mixture-of-Experts Vision-Language Models for Advanced Multimodal Understanding}, 
  author={Zhiyu Wu and Xiaokang Chen and Zizheng Pan and Xingchao Liu and Wen Liu and Damai Dai and Huazuo Gao and Yiyang Ma and Chengyue Wu and Bingxuan Wang and others},
  journal={arXiv preprint arXiv:2412.10302},
  year={2024}
}

@misc{gemini3,
  title = {Gemini 3 Pro: Best for complex tasks and bringing creative concepts to life},
  author = {Gemini Team},
  url = {https://deepmind.google/models/gemini/pro/},
  year = {2025}
}

@article{qwen25vl,
  title={Qwen2.5-VL Technical Report}, 
  author={Shuai Bai and Keqin Chen and Xuejing Liu and Jialin Wang and Wenbin Ge and Sibo Song and Kai Dang and Peng Wang and Shijie Wang and Jun Tang and Humen Zhong and others},
  journal={arXiv preprint arXiv:2502.13923},
  year={2025}
}

@article{liu2023improvedllava,
  title={Improved Baselines with Visual Instruction Tuning}, 
  author={Liu, Haotian and Li, Chunyuan and Li, Yuheng and Lee, Yong Jae},
  journal={arXiv:2310.03744},
  year={2023},
  }

@misc{gemini2,
  title = {Introducing Gemini 2.0: our new AI model for the agentic era},
  author = {Gemini},
  url = {https://blog.google/technology/google-deepmind/google-gemini-ai-update-december-2024/},
  year = {2024}
}

@misc{gemini2think,
  title = {Gemini 2.0 model updates: 2.0 Flash, Flash-Lite, Pro Experimental},
  author = {Gemini},
  url = {https://blog.google/technology/google-deepmind/gemini-model-updates-february-2025/},
  year = {2025}
}

@misc{claude35sonnet,
  title = {Model Card Addendum: Claude 3.5 Haiku and Upgraded Claude 3.5 Sonnet},
  author = {Anthropic},
  url = {https://assets.anthropic.com/m/1cd9d098ac3e6467/original/Claude-3-Model-Card-October-Addendum.pdf},
  year = {2024}
}

@misc{claude4sonnet,
  title = {Introducing Claude 4},
  author = {Anthropic},
  url = {https://www.anthropic.com/news/claude-4},
  year = {2025}
}

@article{gpt4o,
  title={GPT-4o System Card}, 
  author={OpenAI},
  journal={arXiv preprint arXiv:2410.21276},
  year={2024}
}

@misc{gpt4.1,
  title = {Introducing GPT-4.1 in the API},
  author = {OpenAI},
  url = {https://openai.com/index/gpt-4-1/},
  year = {2025}
}

@misc{grok3,
  title = {Grok 3 Beta — The Age of Reasoning Agents},
  author = {xAI},
  url = {https://x.ai/news/grok-3},
  year = {2025}
}

@misc{grok1.5,
  title={Grok-1.5 Vision Preview},
  author={xAI},
  url={https://x.ai/news/grok-1.5v},
  year={2024},
}

@misc{o4-mini,
  title = {OpenAI o3 and o4-mini System Card},
  author = {OpenAI},
  url = {https://openai.com/index/o3-o4-mini-system-card/},
  year = {2025}
}

@misc{glm4v,
  title = {GLM-4V},
  author = {Zhipu.ai},
  url = {https://www.bigmodel.cn/dev/api/normal-model/glm-4v},
  year = {2024}
}

@misc{doubao-1.5-thinking-vision-pro,
  title = {Doubao-1.5-thinking-vision-pro},
  author = {ByteDance Seed},
  url = {https://console.volcengine.com/ark/region:ark+cn-beijing/model/detail?Id=doubao-1-5-thinking-vision-pro},
  year = {2025}
}

@inproceedings{wu2024v,
  title={V*: Guided Visual Search as a Core Mechanism in Multimodal LLMs},
  author={Wu, Penghao and Xie, Saining},
  booktitle={CVPR},
  year={2024}
}

@article{mathvision,
  title={Measuring Multimodal Mathematical Reasoning with MATH-Vision Dataset}, 
  author={Ke Wang and Junting Pan and Weikang Shi and Zimu Lu and Mingjie Zhan and Hongsheng Li},
  journal={arXiv:2402.14804},
  year={2024}
}

@inproceedings{zhang2024mathverse,
  title={Mathverse: Does your multi-modal llm truly see the diagrams in visual math problems?},
  author={Zhang, Renrui and Jiang, Dongzhi and Zhang, Yichi and Lin, Haokun and Guo, Ziyu and Qiu, Pengshuo and Zhou, Aojun and Lu, Pan and Chang, Kai-Wei and Qiao, Yu and others},
  booktitle={ECCV},
  year={2024},
}

@article{xiao2024logicvista,
  title={Logicvista: Multimodal llm logical reasoning benchmark in visual contexts},
  author={Xiao, Yijia and Sun, Edward and Liu, Tianyu and Wang, Wei},
  journal={arXiv preprint arXiv:2407.04973},
  year={2024}
}

@article{xu2025visulogic,
  title={Visulogic: A benchmark for evaluating visual reasoning in multi-modal large language models},
  author={Xu, Weiye and Wang, Jiahao and Wang, Weiyun and Chen, Zhe and Zhou, Wengang and Yang, Aijun and Lu, Lewei and Li, Houqiang and Wang, Xiaohua and Zhu, Xizhou and others},
  journal={arXiv preprint arXiv:2504.15279},
  year={2025}
}

@article{qiao2024we,
  title={We-math: Does your large multimodal model achieve human-like mathematical reasoning?},
  author={Qiao, Runqi and Tan, Qiuna and Dong, Guanting and Wu, Minhui and Sun, Chong and Song, Xiaoshuai and GongQue, Zhuoma and Lei, Shanglin and Wei, Zhe and Zhang, Miaoxuan and others},
  journal={arXiv preprint arXiv:2407.01284},
  year={2024}
}

@article{li2023evaluating,
  title={Evaluating object hallucination in large vision-language models},
  author={Li, Yifan and Du, Yifan and Zhou, Kun and Wang, Jinpeng and Zhao, Wayne Xin and Wen, Ji-Rong},
  journal={arXiv:2305.10355},
  year={2023}
}

@article{chen2024we,
  title={Are We on the Right Way for Evaluating Large Vision-Language Models?},
  author={Chen, Lin and Li, Jinsong and Dong, Xiaoyi and Zhang, Pan and Zang, Yuhang and Chen, Zehui and Duan, Haodong and Wang, Jiaqi and Qiao, Yu and Lin, Dahua and others},
  journal={arXiv:2403.20330},
  year={2024}
}

@article{yu2024mm,
  title={Mm-vet v2: A challenging benchmark to evaluate large multimodal models for integrated capabilities},
  author={Yu, Weihao and Yang, Zhengyuan and Ren, Lingfeng and Li, Linjie and Wang, Jianfeng and Lin, Kevin and Lin, Chung-Ching and Liu, Zicheng and Wang, Lijuan and Wang, Xinchao},
  journal={arXiv preprint arXiv:2408.00765},
  year={2024}
}

@article{liu2024ocrbenchhiddenmysteryocr,
  title={OCRBench: On the Hidden Mystery of OCR in Large Multimodal Models}, 
  author={Yuliang Liu and Zhang Li and Mingxin Huang and Biao Yang and Wenwen Yu and Chunyuan Li and Xucheng Yin and Cheng-lin Liu and Lianwen Jin and Xiang Bai},
  journal={arXiv:2305.07895},
  year={2023}
}

@inproceedings{fu2024blink,
  title={Blink: Multimodal large language models can see but not perceive},
  author={Fu, Xingyu and Hu, Yushi and Li, Bangzheng and Feng, Yu and Wang, Haoyu and Lin, Xudong and Roth, Dan and Smith, Noah A and Ma, Wei-Chiu and Krishna, Ranjay},
  booktitle={ECCV},
  year={2024}
}

@inproceedings{duan2024vlmevalkit,
  title={Vlmevalkit: An open-source toolkit for evaluating large multi-modality models},
  author={Duan, Haodong and Yang, Junming and Qiao, Yuxuan and Fang, Xinyu and Chen, Lin and Liu, Yuan and Dong, Xiaoyi and Zang, Yuhang and Zhang, Pan and Wang, Jiaqi and others},
  booktitle={ACMMM},
  year={2024}
}

@inproceedings{kembhavi2016ai2d,
  title={A diagram is worth a dozen images},
  author={Kembhavi, Aniruddha and Salvato, Mike and Kolve, Eric and Seo, Minjoon and Hajishirzi, Hannaneh and Farhadi, Ali},
  booktitle={ECCV},
  year={2016}
}

@article{li2023seed,
  title={Seed-bench: Benchmarking multimodal llms with generative comprehension},
  author={Li, Bohao and Wang, Rui and Wang, Guangzhi and Ge, Yuying and Ge, Yixiao and Shan, Ying},
  journal={arXiv preprint arXiv:2307.16125},
  year={2023}
}

@article{li2024seedbench2plus,
  title={Seed-bench-2-plus: Benchmarking multimodal large language models with text-rich visual comprehension},
  author={Li, Bohao and Ge, Yuying and Chen, Yi and Ge, Yixiao and Zhang, Ruimao and Shan, Ying},
  journal={arXiv preprint arXiv:2404.16790},
  year={2024}
}

@inproceedings{lu2022learn,
  title={Learn to Explain: Multimodal Reasoning via Thought Chains for Science Question Answering},
  author={Lu, Pan and Mishra, Swaroop and Xia, Tony and Qiu, Liang and Chang, Kai-Wei and Zhu, Song-Chun and Tafjord, Oyvind and Clark, Peter and Ashwin Kalyan},
  booktitle={NeurIPS},
  year={2022}
}

@article{mmtbench,
  title={MMT-Bench: A Comprehensive Multimodal Benchmark for Evaluating Large Vision-Language Models Towards Multitask AGI},
  author={Ying, Kaining and Meng, Fanqing and Wang, Jin and Li, Zhiqian and Lin, Han and Yang, Yue and Zhang, Hao and Zhang, Wenbo and Lin, Yuqi and Liu, Shuo and Lei, Jiayi and Lu, Quanfeng and Chen, Runjian and Xu, Peng and Zhang, Renrui and Zhang, Haozhe and Gao, Peng and Wang, Yali and Qiao, Yu and Luo, Ping and Zhang, Kaipeng and Shao, Wenqi},
  journal={arXiv preprint arXiv:2404.16006},
  year={2024}
}

@article{liu2023mmbench,
  title={MMBench: Is Your Multi-modal Model an All-around Player?},
  author={Liu, Yuan and Duan, Haodong and Zhang, Yuanhan and Li, Bo and Zhang, Songyang and Zhao, Wangbo and Yuan, Yike and Wang, Jiaqi and He, Conghui and Liu, Ziwei and others},
  journal={arXiv preprint arXiv:2307.06281},
  year={2023}
}

@inproceedings{li2023pope,
  title={Evaluating Object Hallucination in Large Vision-Language Models},
  author={Li, Yifan and Du, Yifan and Zhou, Kun and Wang, Jinpeng and Zhao, Wayne Xin and Wen, Ji-Rong},
  booktitle={EMNLP},
  year={2023}
}

@article{zerobench,
  title={ZeroBench: An Impossible Visual Benchmark for Contemporary Large Multimodal Models},
  author={Jonathan Roberts and Mohammad Reza Taesiri and Ansh Sharma and Akash Gupta and Samuel Roberts and Ioana Croitoru and Simion-Vlad Bogolin and Jialu Tang and Florian Langer and Vyas Raina and Vatsal Raina and Hanyi Xiong and Vishaal Udandarao and Jingyi Lu and Shiyang Chen and Sam Purkis and Tianshuo Yan and others},
  journal={arXiv preprint arXiv:2502.09696},
  year={2025}
}

@article{song2025visualpuzzlesdecouplingmultimodalreasoning,
  title={VisualPuzzles: Decoupling Multimodal Reasoning Evaluation from Domain Knowledge},
  author={Yueqi Song and Tianyue Ou and Yibo Kong and Zecheng Li and Graham Neubig and Xiang Yue},
  journal={arXiv preprint arXiv:2504.10342},
  year={2025}
}

@article{internvl3,
  title={InternVL3: Exploring Advanced Training and Test-Time Recipes for Open-Source Multimodal Models},
  author={Jinguo Zhu and Weiyun Wang and Zhe Chen and Zhaoyang Liu and Shenglong Ye and Lixin Gu and Hao Tian and Yuchen Duan and Weijie Su and Jie Shao and Zhangwei Gao and Erfei Cui and Xuehui Wang and Yue Cao and Yangzhou Liu and Xingguang Wei and Hongjie Zhang and Haomin Wang and Weiye Xu and Hao Li and Jiahao Wang and Nianchen Deng and Songze Li  and others},
  journal={arXiv preprint arXiv:2504.10479},
  year={2025}
}

@article{wang20258020rulehighentropyminority,
  title={Beyond the 80/20 Rule: High-Entropy Minority Tokens Drive Effective Reinforcement Learning for LLM Reasoning},
  author={Shenzhi Wang and Le Yu and Chang Gao and Chujie Zheng and Shixuan Liu and Rui Lu and Kai Dang and Xionghui Chen and Jianxin Yang and Zhenru Zhang and Yuqiong Liu and An Yang and Andrew Zhao and Yang Yue and Shiji Song and Bowen Yu and Gao Huang and Junyang Lin},
  journal={arXiv preprint arXiv:2506.01939},
  year={2025}
}

@String(CVPR= {IEEE Conf. Comput. Vis. Pattern Recog.})

@String(ECCV= {Eur. Conf. Comput. Vis.})

@String(ACMMM= {ACM Int. Conf. Multimedia})

@String(ICLR = {Int. Conf. Learn. Represent.})

@String(AAAI = {AAAI})

@String(CVPR  = {CVPR})

@String(ECCV  = {ECCV})

@String(ACMMM = {ACM MM})

@String(ICLR  = {ICLR})
}

\appendix
% \section{Appendix}

\clearpage
\section{Dataset Statistics}
\label{app:statistic}
To construct our dataset, we leveraged the 1,434 high-quality metaphorical images from II-Bench \citep{liu2024iibenchimageimplicationunderstanding}. II-Bench encompasses images from six distinct domains: Life, Art, Society, Psychology, Environment and Others. It features a diverse array of image types, including Illustrations, Memes, Posters, Multi-panel Comics, Single-panel Comics, Logos and Paintings. 
We manually construct the TFQ-Data-Lite and TFQ-Bench-Lite by selecting 50-100 high-quality, diverse and representative images. The general statistic is in Table~\ref{tab:dataset_statistics_1} \ref{tab:dataset_statistics_2} \ref{tab:dataset_statistics_3}.

\begin{table}[h!]
    \centering
    \footnotesize
        \begin{tabular}{@{}ll@{}}
            \toprule
            \multicolumn{2}{l}{\textbf{Statistics of TFQ-Data \& TFQ-Bench Images}} \\ \cmidrule(r){1-2}
            Life        & 516 (42.2\%)   \\
            Art      & 70 (5.7\%)   \\
            Society         & 408 (33.4\%)\\
            Psychology    & 127 (10.4\%)   \\
            Environment            & 44 (3.6\%) \\ 
            Other                & 57 (4.7\%)       \\     \midrule
           
            Positive        & 169 (13.8\%)      \\
            Neutral          & 702 (57.5\%)       \\
            Negative         & 351 (28.7\%)        \\    \midrule
           
            Illustration                 & 374 (28.7\%)  \\
            Meme                 & 269 (20.6\%) \\
            Poster                 & 111 (8.5\%) \\
            Multi-panel Comic                & 311 (23.9\%)  \\
            Single-panel Comic                & 90 (6.9\%) \\
            Logo                & 59 (4.5\%) \\
            Painting      & 89 (6.8\%)    \\ \bottomrule
        \end{tabular}
    % \vspace{0.3cm}
    \caption{General statistics of the TFQ-Data and TFQ-Bench.}
    \label{tab:dataset_statistics_1}
    \vspace{-0.3cm}
\end{table}

\begin{table}[h!]
    \centering
    \footnotesize
        \begin{tabular}{@{}ll@{}}
            \toprule
            \multicolumn{2}{l}{\textbf{Statistics of TFQ-Data-Lite Images}} \\ \cmidrule(r){1-2}
            Life     & 39 (39\%) \\
            Society      & 23 (23\%)   \\
            Psychology  & 19 (19\%) \\ 
            Art         & 12 (12\%)   \\
            Environment   & 6 (6\%) \\ 
            Others  & 1 (1\%) \\     
            \midrule
            Multi-panel Comic  & 33 (28.7\%)  \\
            Meme               & 22 (19.1\%) \\
            Illustration       & 20 (17.4\%) \\
            Poster             & 17 (14.8\%) \\
            Logo           & 15 (13.0\%)  \\              
            Single-panel Comic & 7 (6.1\%) \\
            Painting           & 1 (0.9\%)  \\  
            \bottomrule
        \end{tabular}
    % \vspace{0.3cm}
    \caption{General statistics of the TFQ-Data-Lite.}
    \label{tab:dataset_statistics_2}
    \vspace{-0.3cm}
\end{table}

\begin{table}[h!]
    \centering
    \footnotesize
        \begin{tabular}{@{}ll@{}}
            \toprule
            \multicolumn{2}{l}{\textbf{Statistics of TFQ-Bench-Lite Images}} \\ \cmidrule(r){1-2}
            Society     & 21 (42\%) \\
            Life        & 16 (32\%)   \\
            Art         & 6 (2\%)   \\
            Psychology  & 4 (8\%) \\ 
            Others  & 3 (6\%) \\     
            \midrule
            Multi-panel Comic  & 16 (32\%)  \\
            Single-panel Comic & 9 (18\%) \\
            Illustration       & 5 (10\%) \\
            Meme               & 5 (10\%) \\
            Poster             & 5 (10\%) \\
            Painting           & 5 (10\%)  \\  
            Logo           & 5 (10\%)  \\  
            \bottomrule
        \end{tabular}
    % \vspace{0.3cm}
    \caption{General statistics of the TFQ-Bench-Lite.}
    \label{tab:dataset_statistics_3}
        \vspace{-0.3cm}
\end{table}

\section{Experiment Setup}
\label{app:experiment_details}
\textbf{Parameter Details.} We set the model temperature as 0.5 and top\_p as 0.9 in TFQ and MCQ experiments, and temperature as 0.7 and top\_p as 0.9 in OSQ experiments. Additionally, we set the evaluation model GPT-4o temperature as 0 and evaluate more than three times to get the average score in OSQ experiments. And the average human-model scoring consistency reached 96.5\% on OSQ  \citep{zhang2025letandroidsdream}.
All experiments are conducted on NVIDIA A800 and H200 GPUs.

\textbf{Main Experiment.} To comprehensively compare with the MetaphorStar family, we carefully select a diverse range of MLLMs, encompassing both open-source and closed-source models, with the aim of covering a wide spectrum of model characteristics and scales. These models span parameter sizes from 7B to 300B, ensuring that models of varying complexity and capability are thoroughly assessed. In selecting the models, we focus on the following key aspects: 1) General and Reasoning models, 2) Open-Source and Closed-Source models, and 3) model parameter scaling law. 

The high-level bench (EN) \citep{zhang2025letandroidsdream}, which is manually constructed by 50 high-quality, diverse, and representative English images from varied image types like illustrations and comics, featuring Multiple-Choice Question (MCQ) and Open-Style Question (OSQ). And the average human-model scoring consistency reached 96.5\% on OSQ \citep{zhang2025letandroidsdream}.

\textbf{Generalization Experiment.} We mainly select two categories of benchmarks — Reasoning and Understanding.

% Reasoning
We provide a comprehensive review of benchmarks specifically designed to assess various facets of MLLM reasoning capabilities, which are critical for their deployment in environments requiring complex decision-making.
Therefore, we select MathVision~\citep{MathVision}, MathVerse~\citep{zhang2024mathverse}, WeMath~\citep{qiao2024we}, LogicVista~\citep{xiao2024logicvista}, VisuLogic~\citep{xu2025visulogic}, VisualPuzzles~\citep{song2025visualpuzzlesdecouplingmultimodalreasoning}, V*~\citep{wu2024v}, ZeroBench~\citep{zerobench}, and MMMU~\cite{yue2023mmmu} to verify the model's reasoning ability.

% Understanding
We revisit multimodal understanding benchmarks designed to assess MLLMs’ ability to perceive and comprehend information presented in various formats, such as text and images. These benchmarks are crucial for fine-tuning MLLMs, ensuring their robustness and generalization in real-world applications.
These benchmarks include SEEDBench~\cite{li2023seed}, SEED-2-Plus~\cite{li2024seedbench2plus}, MMBench (English)~\cite{liu2023mmbench}, MMBench v1.1 (English)~\cite{liu2023mmbench}, MMStar~\citep{chen2024we}, OCRBench~\citep{liu2024ocrbenchhiddenmysteryocr}, AI2D~\cite{kembhavi2016ai2d}, ScienceQA~\citep{lu2022learn}, POPE~\cite{li2023pope}, MMT-Bench~\cite{mmtbench}, RealWorld QA~\citep{grok1.5}, BLINK~\citep{fu2024blink}, HallusionBench~\citep{li2023evaluating}, and MMVet Hard~\citep{yu2024mm}.

\end{document}